\documentclass{article}

     \PassOptionsToPackage{numbers, compress}{natbib}

   \usepackage[preprint]{neurips_2025}

\usepackage[utf8]{inputenc} 
\usepackage[T1]{fontenc}    
\usepackage{url}            
\usepackage{booktabs}       
\usepackage{amsfonts}       
\usepackage{nicefrac}       
\usepackage{microtype}      
\usepackage{xcolor}         
\usepackage[pdftex]{graphicx}
\usepackage{epstopdf}
\usepackage{multirow}
\usepackage{colortbl}
\usepackage[pagebackref=true,breaklinks=true,letterpaper=true,colorlinks,bookmarks=false]{hyperref}
\usepackage{subcaption}
\usepackage[namelimits]{amsmath} 
\usepackage{amssymb}            
\usepackage{mathrsfs}            
\usepackage{bm}
\usepackage{wrapfig}
\usepackage{bbm}
\usepackage{cleveref}
\usepackage{algorithm}
\usepackage{algorithmic}
\usepackage{marvosym}
\usepackage{inconsolata}

\makeatletter
\newcommand*\bigcdot{\mathpalette\bigcdot@{.5}}
\newcommand*\bigcdot@[2]{\mathbin{\vcenter{\hbox{\scalebox{#2}{$\m@th#1\bullet$}}}}}
\makeatother

\definecolor{c2}{HTML}{FBD9BD}
\definecolor{c3}{HTML}{fe793d}
\definecolor{c4}{HTML}{eedeb0}
\definecolor{pp}{HTML}{BC7FCD}
\definecolor{bb}{HTML}{CDE8E5}
\definecolor{rouse}{rgb}{0.981,0.961,0.941}

\crefname{section}{Sec.}{Secs.}
\Crefname{section}{Section}{Sections}
\Crefname{table}{Tab.}{Tabs.}
\Crefname{table}{Table}{Tables}
\crefname{figure}{Fig.}{Figs.}
\crefname{equation}{Eq.}{Eqs.}

\title{\textbf{UnfoldIR}: Rethinking Deep Unfolding Network in \\ Illumination Degradation Image Restoration}

\author{
 Chunming He$^{1}$\,,
 Rihan Zhang$^{1}$\,,
 Fengyang Xiao$^{1}$\,, 
  \textbf{Chengyu Fang}$^2$\,, \\
  \textbf{Longxiang Tang}$^2$\,, 
        \textbf{Yulun Zhang}$^{3}$\,, 
	\textbf{Sina Farsiu$^{1,*}$}\\
	$^1$Duke University, ~$^3$Tsinghua University, ~ $^3$Shanghai Jiao Tong University
 }

\begin{document}

\maketitle

\begin{abstract} \label{abstract}
Deep unfolding networks (DUNs) are widely employed in illumination degradation image restoration (IDIR) to merge the interpretability of model-based approaches with the generalization of learning-based methods. 
However, the performance of DUN-based methods remains considerably inferior to that of state-of-the-art IDIR solvers. Our investigation indicates that this limitation does not stem from structural shortcomings of DUNs but rather from the limited exploration of the unfolding structure, particularly for (1) constructing task-specific restoration models, (2) integrating advanced network architectures, and (3) designing DUN-specific loss functions.
To address these issues, we propose a novel DUN-based method, UnfoldIR, for IDIR tasks. 
UnfoldIR first introduces a new IDIR model with dedicated regularization terms for smoothing illumination and enhancing texture. 
We unfold the iterative optimized solution of this model into a multistage network, with each stage comprising a reflectance-assisted illumination correction (RAIC) module and an illumination-guided reflectance enhancement (IGRE) module. 
RAIC employs a visual state space (VSS) to extract non-local features, enforcing illumination smoothness, while IGRE introduces a frequency-aware VSS to globally align similar textures, enabling mildly degraded regions to guide the enhancement of details in more severely degraded areas. This suppresses noise while enhancing details.
Furthermore, given the multistage structure, we propose an inter-stage information consistent loss to maintain network stability in the final stages. This loss contributes to structural preservation and sustains the model's performance even in unsupervised settings.
Experiments verify our effectiveness across 5 IDIR tasks and 3 downstream problems. Besides, our analysis of the intrinsic mechanisms of DUNs provides valuable insights for future research. Code will be released.

\end{abstract}

\setlength{\abovedisplayskip}{2pt}
\setlength{\belowdisplayskip}{2pt}
\section{Introduction} \label{introduction} 

\begin{figure*}[h]
\setlength{\abovecaptionskip}{0cm} 	\setlength{\belowcaptionskip}{0cm}
\includegraphics[width=1\linewidth]{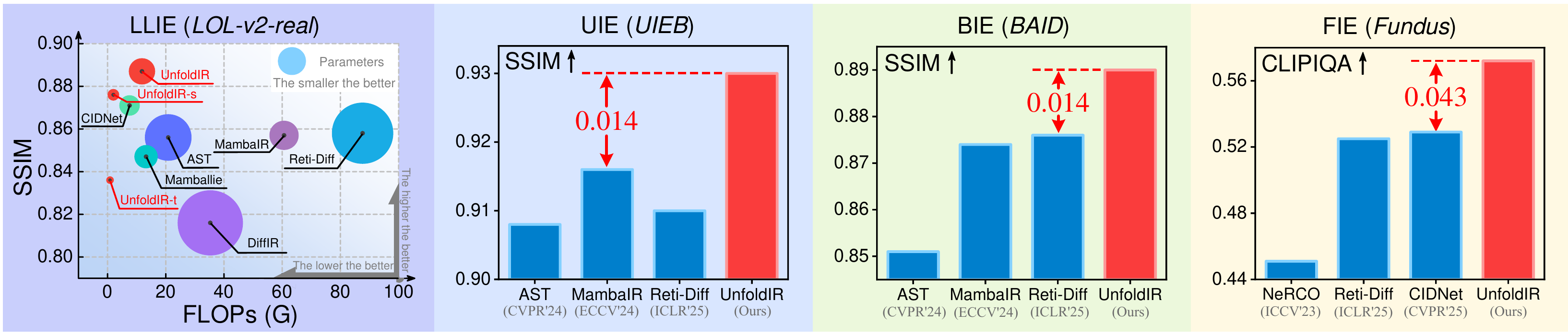}
	\captionof{figure}{Results on IDIR tasks with commonly used datasets and metrics. In low-light image enhancement (LLIE), we compare our methods, UnfoldIR and its two lightweight variants (UnfoldIR-t and UnfoldIR-s), with cutting-edge methods. Our advancement is also proven by underwater image enhancement (UIE), backlit image enhancement (BIE), and fundus image enhancement (FIE).}
    \label{fig:Intro}
	\vspace{-9mm}
\end{figure*}

Illumination degradation image restoration (IDIR)~\cite{he2025reti,guo2023underwater,liang2023iterative,he2023hqg}, with representative tasks listed in~\cref{fig:Intro}, refers to a set of challenging restoration tasks in which images suffer from the adverse effects of degraded illumination, such as low contrast and non-uniform noise.
By addressing illumination degradation, the restored data are expected to exhibit enhanced details and improved fidelity, thereby facilitating downstream tasks like nighttime object detection.
To achieve this, substantial algorithms have been proposed~\cite{fu2016weighted,ueng1995gamma,jiang2021enlightengan,yi2023diff}, mainly including
model-based and deep learning-based methods.

Model-based methods~\cite{fu2016weighted,ueng1995gamma} rely on manually designed restoration rules, which afford clear interpretability but suffer from limited generalization. In contrast, learning-based strategies~\cite{jiang2021enlightengan,yi2023diff}, which benefit from end-to-end training, have achieved significant success in IDIR tasks due to their improved generalization, although they remain less interpretable.
To unify the merits of model-based and learning-based methods, deep unfolding networks (DUNs) have been proposed for IDIR. 
These methods~\cite{wu2022uretinex,zheng2023empowering} unfold the iterative optimized solution of a task-specific model—typically one based on the Retinex model—into a multi-stage network, converting fixed parameters into learnable ones.

Notably, the performance of existing DUN-based methods is lower than that of state-of-the-art IDIR solvers, which challenges the applicability of DUN-based frameworks. This limitation does not reflect structural flaws in DUNs but rather indicates that the potential of the unfolding structure has not been fully realized—especially for \textit{(1) constructing restoration models with task-specific regularization terms during the modeling phase, (2) integrating advanced networks within the unfolding process}, and \textit{(3) designing loss functions that align with DUN structures during training time}.

To address these issues, we propose a novel DUN-based method, UnfoldIR, for IDIR tasks. First, UnfoldIR introduces a new IDIR model (IDIRM) based on the Retinex theory, which decomposes the input image into illumination and reflectance components. 
IDIRM incorporates dedicated regularization terms for the illumination component to enforce smoothness and for the reflectance component to suppress noise and enhance texture. We then use the proximal gradient algorithm to optimize the model and unfold the iterative solution into a multi-stage network architecture, UnfoldIR.

In UnfoldIR, each stage comprises two modules: the reflectance-assisted illumination correction (RAIC) module and the illumination-guided reflectance enhancement (IGRE) module, which iteratively optimize the illumination and reflectance components.
The RAIC module employs a visual state space (VSS) to extract non-local features from the illumination component, effectively reducing color distortion and maintaining global color consistency. 
In addition to mitigating the inherent imaging noise in the reflectance, the IGRE module must also suppress noise introduced during the illumination recovery process. To achieve this, we introduce a frequency-aware VSS that globally aligns similar texture details using the estimated illumination as a condition. This mechanism allows lightly degraded regions to guide the enhancement of details in more severely degraded areas. Moreover, we integrate the VSS into a high-order ordinary differential equation framework—specifically, a second-order Runge-Kutta method—to accurately suppress noise and enhance image details.

Additionally, given our multi-stage structure, we propose an inter-stage information consistent (ISIC) loss. The ISIC loss is designed to maintain stability in the final stages so that small changes in illumination do not compromise essential reflectance details in the restored image, and vice versa.
This contributes to structural preservation and distortion reduction.  Moreover, since the ISIC loss serves as a framework-specific self-constraint, it can also be employed in unsupervised settings.

This paper also explores the intrinsic advantages of DUNs and investigates the deployment for image restoration in~\cref{discussion}, aiming to guide the design of future DUN-based methods in this domain.

Our contributions are summarized as follows:

(1) We propose UnfoldIR, a novel deep unfolding network designed for IDIR tasks, which incorporates a new IDIR model designed to enforce illumination smoothness, suppress noise, and enhance texture.


(2) UnfoldIR introduces two new modules—RAIC and IGRE—specifically tailored to the illumination and reflectance components. The RAIC module enforces illumination smoothness, while the IGRE module enhances structural details and suppresses undesired noise in the reflectance component.


(3) We propose a self-supervised ISIC loss for structural preservation and distortion elimination, maintaining the model’s performance even in unsupervised settings.

(4) Experiments validate our effectiveness across 5 IDIR tasks and 3 downstream problems. 
Besides, our analysis of DUN’s internal mechanisms provides insights to guide future DUN-based methods.

\begin{figure*}[th]
\setlength{\abovecaptionskip}{0.05cm}
	\centering
	\includegraphics[width=\linewidth]{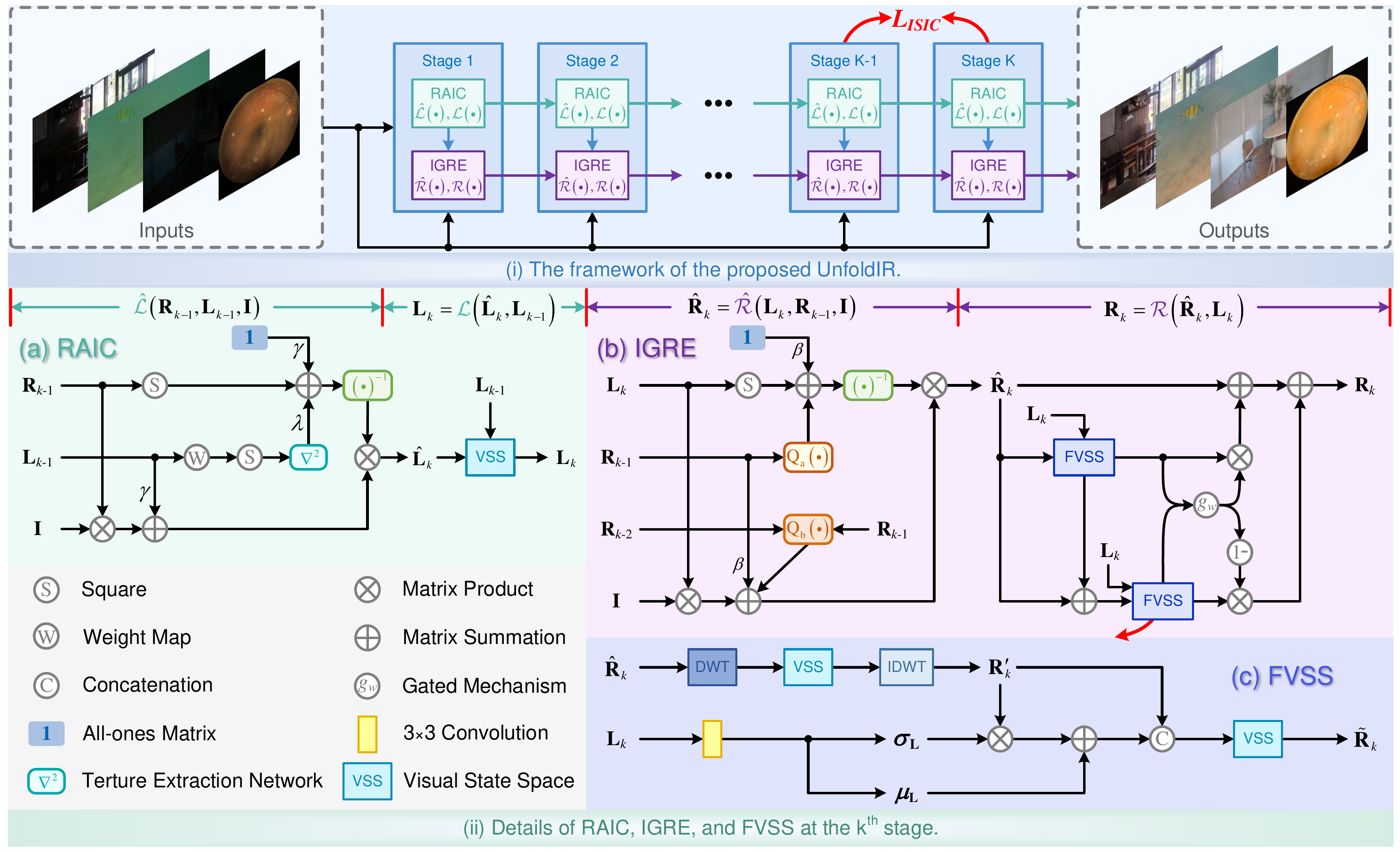} 
	\caption{Framework of our UnfoldIR. The network connection in $\hat{\mathcal{L}}(\bigcdot)$ and $\hat{\mathcal{R}}(\bigcdot)$ are derived strictly based on mathematical principles, thus enhancing interpretability. For clarity, we replace certain redundant details with $\mathbf{Q}_{\mathbf{a}}$ and $\mathbf{Q}_{\mathbf{b}}$, and present $\hat{\mathcal{R}}(\bigcdot)$ according to \cref{Eq:OptimizeR3}.    }
	\label{fig:framework}
	\vspace{-3.5mm}
\end{figure*}
\section{Related Works}\vspace{-1.5mm}
\noindent\textbf{Illumination Degradation Image Restoration}. IDIR has progressed from traditional methods to deep learning-based solutions~\cite{cheng2004simple,he2019image,huang2012efficient,liu2021retinex}.
Among them, Retinex theory-based methods play a significant role in decomposing inputs into reflectance and illumination components. 
Traditional ones typically introduce explicit Retinex priors to constrain Retinex maps~\cite{fu2016weighted,li2018structure}, offering interpretability but limited generalization. 
In contrast, learning-based methods, although lacking interpretability, achieve SOTA performance due to their powerful feature extractors. 
For example, DIE~\cite{wang2019underexposed} integrated Retinex cues with a one-stage framework to address color distortion. 
Reti-diff~\cite{he2025reti} added estimated Retinex priors into a transformer for efficient inference. 
Besides, DUN, combining the merits of traditional and learning-based methods, has recently been introduced into IDIR, with the potential to achieve superior results. However, existing DUNs~\cite{wu2022uretinex,zheng2023empowering} still underperform compared to current SOTAs.

\noindent\textbf{Deep Unfolding Network}. DUNs, such as 
DM-Fusion~\cite{xu2023dm} and DeRUN~\cite{he2023degradation}, aim to unfold 
iterative optimization algorithms into deep networks, converting traditionally fixed parameters and operators into learnable ones. 
Several DUN-based methods have been introduced in IDIR. 
For instance, Uretinex-Net~\cite{wu2022uretinex} unfolded the Retinex model into a multi-stage network, mitigating noise interference. 
CUE~\cite{zheng2023empowering} leveraged a masked autoencoder-based loss function to supervise an unfolded network for learning customized priors. 
However, these methods still fall notably short of the existing SOTAs. 
Our analysis indicates that this limitation does not stem from inherent structural deficiencies in DUNs but rather from the insufficient exploration of the unfolding structures.
To address this, we propose UnfoldIR, a novel DUN-based framework tailored for IDIR tasks. 
Specifically, UnfoldIR constructs a task-specific restoration model (IDIRM), incorporates advanced VSS structures to form RAIC and IGRE modules, and introduces a DUN-specific ISIC loss to maintain network stability. With these enhancements, our UnfoldIR achieves leading performance across five IDIR tasks.

\section{Methodology} \label{methodology}

\subsection{IDIR Model}
According to Retinex theory, an illumination-degraded image $\mathbf{I}$ can be decomposed into its reflectance image $\mathbf{R}$ and illumination map $\mathbf{L}$ using the Hadamard product $\odot$, formulated as
\begin{equation}\label{Eq:Retinex}
    \mathbf{I}=\mathbf{R} \odot \mathbf{L}=(\mathbf{R}_{HQ}+\hat{\mathbf{R}}) \odot (\mathbf{L}_{HQ}+\hat{\mathbf{L}}), 
\end{equation}
where $\mathbf{R}_{HQ}$ and $\mathbf{L}_{HQ}$ denote the latent high-quality reflectance and illumination components, and $\hat{\mathbf{R}}$ and $\hat{\mathbf{L}}$ represent perturbations corresponding to textural degradation and color distortion. To reduce these perturbations, we restore $\mathbf{R}$ and $\mathbf{L}$ by optimizing the following objective function:
\begin{equation}\label{Eq:BasicModel}
    L(\mathbf{R},\mathbf{L}) = \frac{1}{2}\|\mathbf{I}-\mathbf{R} \odot \mathbf{L}\|_2^2 + \beta\varphi(\mathbf{R}) + \gamma \phi(\mathbf{L}), 
\end{equation}
where $\|\bigcdot\|_2$ is a $\ell_2$-norm. $\beta$ and $\gamma$ are trade-off parameters. 
The regularization terms $\varphi(\mathbf{R})$ and $\phi(\mathbf{L})$ will be implicitly learned by deep networks to effectively suppress the perturbations.
In addition to these implicit terms, we propose incorporating explicit constraints into the reflectance $\mathbf{R}$ to suppress imaging noise and enhance texture details, and into the illumination $\mathbf{L}$ to reduce color distortion based on their degradation characteristics. Consequently, the objective function becomes:
\begin{equation}\label{Eq:FinalModel}
    L(\mathbf{R},\mathbf{L}) = \frac{1}{2}\|\mathbf{I}-\mathbf{R} \odot \mathbf{L}\|_2^2 + \beta\varphi(\mathbf{R}) + \gamma \phi(\mathbf{L})+\mu \mathcal{S}\!\left(\mathcal{A}(\mathbf{I})-\mathcal{A}(\mathbf{R})\right)+\frac{\lambda}{2}\|\mathbf{w}\odot\nabla \mathbf{L}\|_2^2,
\end{equation}
where $\mu$ and $\lambda$ are trade-off parameters. 
Inspired by the Perona-Malik algorithm~\cite{perona1990scale}, we introduce $\mathcal{A}(\bigcdot)$ and use $\mathcal{S}(\bigcdot)$, an $\ell_1$-norm, to enforce texture distribution consistency between $\mathbf{I}$ and $\mathbf{R}$.
At pixel location $i$, $\mathcal{A}(\mathbf{I}_i)=\frac{1}{\tilde{\eta}_i}\sum_{j\in \eta_i}c(|\nabla \mathbf{I}_{i,j}|)\nabla \mathbf{I}_{i,j}$, 
where $\eta_i$ is the set of neighboring pixels around pixel $i$, and $\tilde{\eta}_i$ is the number of neighboring pixels, set to 9. $|\bigcdot|$ means gradient magnitude and $\nabla$ is the gradient operator, such as Sobel operator~\cite{sobel19683x3}. 
The diffusion function $c(|\nabla \mathbf{I}_{i,j}|)$ is: $c(|\nabla \mathbf{I}_{i,j}|)=\text{exp}[-(\frac{|\nabla \mathbf{I}_{i,j}|}{s})^2]$, where $s$ is a constant controlling the sensitivity of the diffusion process and is set as a learnable parameter in UnfoldIR. 
A small $|\nabla \mathbf{I}_{i,j}|$ indicates a relatively smooth region, thereby intensifying the diffusion process and suppressing noise. 
Conversely, a large one implies the presence of important texture details, weakening the diffusion effect and enhancing texture preservation.

For the illumination map $\mathbf{L}$, we propose a local smoothing constraint to mitigate color distortion. $\mathbf{w}$ is a gradient-aware weighting matrix and $\mathbf{w}=\frac{1}{{\text{exp}}|\nabla \mathbf{L}|}$. This weighting matrix effectively preserves significant structural information while preventing over-smoothing.


\subsection{UnfoldIR}
\subsubsection{Model Optimization}
We utilize the proximal gradient algorithm~\cite{fang2024real} to optimize~\cref{Eq:FinalModel}, progressively suppressing the perturbations $\hat{\mathbf{R}}$ and $\hat{\mathbf{L}}$, and ultimately obtaining the optimal Retinex components $\mathbf{R}^*$ and $\mathbf{L}^*$:
\begin{equation}
   \{\mathbf{R}^*,\mathbf{L}^*\}= \arg \min_{\mathbf{R},\mathbf{L}}L(\mathbf{R},\mathbf{L}).
\end{equation}
The optimization process involves alternating updates of $\mathbf{L}$ and $\mathbf{R}$ over iterations. 
In the following, we select the $k^{th}$ iteration ($1\leq k\leq K$) to present the alternative solution process. 

\noindent\textbf{Optimizing $\mathbf{L}_k$}. The optimization function is partitioned to update the illumination map $\mathbf{L}_k$:
\begin{equation}\label{Eq:OptimizeL}
   \mathbf{L}_k = \arg \min_{\mathbf{L}}L(\mathbf{R}_{k-1},\mathbf{L}) = \arg \min_{\mathbf{L}}\frac{1}{2}\|\mathbf{I}-\mathbf{R}_{k-1} \odot \mathbf{L}\|_2^2 + \gamma \phi(\mathbf{L})+\frac{\lambda}{2}\|\mathbf{w}_{k}\odot\nabla \mathbf{L}\|_2^2.
\end{equation}
The solution comprises two terms, that is, the gradient descent term and the proximal term. By introducing an auxiliary variable $\hat{\mathbf{L}}_k$, the two terms can be formulated as
\begin{equation} \label{Eq:OptimizeL1}
   \hat{\mathbf{L}}_k = \frac{1}{2}\|\mathbf{I}-\mathbf{R}_{k-1} \odot \hat{\mathbf{L}}\|_2^2 + \frac{\gamma}{2} \|\hat{\mathbf{L}}-\mathbf{L}_{k-1}\|^2_2+\frac{\lambda}{2}\|\mathbf{w}_k
   \odot\nabla \hat{\mathbf{L}}\|_2^2,
\end{equation}
\begin{equation} \label{Eq:OptimizeL2}
   \mathbf{L}_k = \text{prox}_\phi(\mathbf{R}_{k-1}, \hat{\mathbf{L}}_k),
\end{equation}
where $\mathbf{L}_0$ and $\mathbf{R}_0$ are initialized following Uretinex-Net~\cite{wu2022uretinex} and $\mathbf{w}_k$ is constructed based on $\mathbf{L}_{k-1}$. \cref{Eq:OptimizeL1} can be solved directly by equating its derivative to zero, while \cref{Eq:OptimizeL2} will be replaced by a deep network in~\cref{Sec:DUM}.
The closed-form solution of $\hat{\mathbf{L}}_k$ is
\begin{equation} \label{Eq:OptimizeL3}
   \hat{\mathbf{L}}_k = \left(\mathbf{R}_{k-1}^2 + \lambda\mathbf{w}_k^2\nabla^2+\gamma\mathbf{1}\right)^{-1}\left(\mathbf{R}_{k-1}\mathbf{I}+\gamma \mathbf{L}_{k-1}\right),
\end{equation}
where $\mathbf{1}$ is an all-ones matrix.

\noindent\textbf{Optimizing $\mathbf{R}_k$}. The optimization function of $\mathbf{R}_k$ is formulated as:
\begin{equation} \label{Eq:OptimizeR}
    \mathbf{R}_k=\arg \min_{\mathbf{R}}L(\mathbf{R},\mathbf{L}_k) = \arg \min_{\mathbf{R}}\frac{1}{2}\|\mathbf{I}-\mathbf{R} \odot \mathbf{L}_k\|_2^2 + \beta\varphi(\mathbf{R}) + \mu \mathcal{S} (\mathcal{A}(\mathbf{I})-\mathcal{A}({\mathbf{R}})).
\end{equation}
Same as the optimization rule for $\mathbf{L}_k$, the gradient descent term and the proximal term are defined as:
\begin{equation} \label{Eq:OptimizeR1}
    \hat{\mathbf{R}}_k= \frac{1}{2}\|\mathbf{I}-\hat{\mathbf{R}} \odot \mathbf{L}_k\|_2^2 + \frac{\beta}{2}\|\hat{\mathbf{R}}-\mathbf{R}_{k-1}\|_2^2 + \mu \mathcal{S} (\mathcal{A}(\mathbf{I})-\mathcal{A}(\hat{\mathbf{R}})),
\end{equation}
\begin{equation} \label{Eq:OptimizeR2}
    \mathbf{R}_k=\text{prox}_\varphi(\hat{\mathbf{R}}_k,\mathbf{L}_k).
\end{equation}
The closed-form solution of $\hat{\mathbf{R}}_k$ can be acquired similarly:
\begin{equation} \label{Eq:OptimizeR3}
    \hat{\mathbf{R}}_k= (\mathbf{L}_k^2+\beta \mathbf{1}+\mathbf{Q}_\mathbf{a})^{-1}(\mathbf{L}_k\mathbf{I}+\beta\mathbf{R}_{k-1}+\mathbf{Q}_\mathbf{b})  ,
\end{equation}
where $\mathbf{Q}_\mathbf{a}=\frac{\mu L_\mathcal{S}}{\tilde{\eta}_i^2}\sum_{j\in \eta_i}c(|\nabla {\mathbf{R}}_{k-1}|)^2\nabla^2$, 
$\mathbf{Q}_\mathbf{b}=\frac{\mu L_\mathcal{S}}{\tilde{\eta}_i^2}\sum_{j\in \eta_i} c(|\nabla {\mathbf{R}}_{k-1}|)c(|\nabla {\mathbf{R}}_{k-2}|)\nabla^2+ \frac{\mu }{\tilde{\eta}_i }$ $\sum_{j\in \eta_i}c(|\nabla {\mathbf{R}}_{k-1}|) \nabla \mathcal{S}_L(\frac{1}{\tilde{\eta}_i }\sum_{j\in \eta_i} c(|\nabla {\mathbf{I}}|)\nabla {\mathbf{I}} - c(|\nabla  {\mathbf{R}}_{k-2}|)\nabla {\mathbf{R}}_{k-2})$.
Redundant subscripts are omitted.
$L_\mathcal{S}$ is the Lipschitz constant. $ \mathcal{S}_L(\bigcdot)$ is the Lipschitz continuous gradient function of $\mathcal{S}(\bigcdot)$.

\subsubsection{Deep Unfolding Mechanism}\label{Sec:DUM}
We unfold the iterative solutions into a multi-stage network, UnfoldIR, with each step corresponding to a stage. As shown in~\cref{fig:framework}, each stage has two modules: the reflectance-assisted illumination correction (RAIC) and illumination-guided reflectance enhancement (IGRE) modules. 

\noindent\textbf{RAIC}. RAIC, derived from~\cref{Eq:OptimizeL2,Eq:OptimizeL3}, uses $\hat{\mathcal{L}}(\bigcdot)$ and $\mathcal{L}(\bigcdot)$ to compute the optimized result $\hat{\mathbf{L}}$ and the refined illumination map $\mathbf{L}$, respectively. Given $\mathbf{R}_{k-1}$ and $\mathbf{L}_{k-1}$, we define $\hat{\mathbf{L}}_k$ as follows:
\begin{equation} \label{Eq:RAIC1}
   \hat{\mathbf{L}}_k = \hat{\mathcal{L}}(\mathbf{R}_{k-1},\mathbf{L}_{k-1},\mathbf{I}) = \left(\mathbf{R}_{k-1}^2 + \lambda\mathbf{w}_k^2\nabla^2+\gamma\mathbf{1}\right)^{-1}\left(\mathbf{R}_{k-1}\mathbf{I}+\gamma \mathbf{L}_{k-1}\right),
\end{equation}
\cref{Eq:RAIC1} retains the same formulation as~\cref{Eq:OptimizeL3}, with originally fixed parameters to be learnable.

To refine the estimated illumination map $\hat{\mathbf{L}}_k$, we introduce a visual state space (VSS) module~\cite{liu2024vmamba}, denoted as $VSS(\bigcdot)$, to extract non-local features. This enables the network to adjust global illumination and promotes uniform lighting across the image. The design is particularly effective under backlit conditions, where the illumination often exhibits region-level heterogeneity. Under this formulation, the refined illumination map $\mathbf{L}_{k}$ is given by:
\begin{equation} \label{Eq:RAIC2}
   {\mathbf{L}}_k =  \mathcal{L}(\hat{\mathbf{L}}_{k}, {\mathbf{L}}_{k-1}) = VSS(\hat{\mathbf{L}}_{k}, {\mathbf{L}}_{k-1}).
\end{equation}
\noindent\textbf{IGRE}. in IGRE, the calculation of $\hat{\mathbf{R}}_k$ relies on $\hat{\mathcal{B}}(\bigcdot)$, similar to~\cref{Eq:OptimizeR3} but with fixed parameters, including $\mathcal{S}_L(\bigcdot)$, made learnable. Given $\mathbf{L}_{k}$ and $\mathbf{R}_{k-1}$, $\hat{\mathbf{R}}_k$ can be calculated as
\begin{equation} \label{Eq:IGRE1}
    \hat{\mathbf{R}}_k= \hat{\mathcal{R}}(\mathbf{L}_{k}, \mathbf{R}_{k-1}, \mathbf{R}_{k-2}, \mathbf{I})= (\mathbf{L}_k^2+\beta \mathbf{1}+\mathbf{Q}_\mathbf{a})^{-1}(\mathbf{L}_k\mathbf{I}+\beta\mathbf{R}_{k-1}+\mathbf{Q}_\mathbf{b}).
\end{equation}
Refining the reflectance component $\hat{\mathbf{R}}_k$ is inherently more complex. In addition to mitigating intrinsic imaging noise and recovering weakened texture details—both common in illumination-degraded scenarios—the refinement process must also account for interference introduced by the optimized illumination component $\mathbf{L}$. To address these issues, we begin by unrolling~\cref{Eq:Retinex}, resulting in:
\begin{equation}\label{Eq:IGRE2}
\mathbf{I}=\mathbf{R}_{HQ}\odot\mathbf{L}_{HQ}+\mathbf{R}_{HQ}\odot\hat{\mathbf{L}}+\hat{\mathbf{R}}\odot(\mathbf{L}_{HQ}+\hat{\mathbf{L}}).
\end{equation}
In the third term of~\cref{Eq:IGRE2}, the optimization of $\mathbf{L}$—that is, the light-up process—aims to suppress the perturbation $\hat{\mathbf{L}}$, but it can also amplify noise hidden in the dark scenes. Therefore, the optimization of $\mathbf{R}$ must concurrently address noise suppression during the illumination recovery phase.

To this end, we introduce a frequency-aware VSS (FVSS) module, $FVSS(\bigcdot)$, which conditions on the estimated illumination map $\mathbf{L}_k$. The FVSS module is designed to globally align similar texture patterns and enables lightly degraded regions to guide the enhancement of details in more severely degraded areas under illumination-aware guidance. 
The preliminary refined result $\tilde{\mathbf{R}}_k$ is defined as
\begin{equation} \label{Eq:IGRE3}
    \widetilde{\mathbf{R}}_k= FVSS(\hat{\mathbf{R}}_k,{\mathbf{L}}_k)=VSS(conca(\mathbf{R}'_k,\bm{\sigma}_\mathbf{L}\mathbf{R}'_k+\bm{\mu}_\mathbf{L})), 
\end{equation}
\begin{equation} \label{Eq:IGRE4}
 \mathbf{R}'_k=IDWT(VSS(DWT(\hat{\mathbf{R}}_k))), \bm{\sigma}_\mathbf{L}=conv3_{\bm{\sigma}}(\mathbf{L}_k), 
 \bm{\mu}_\mathbf{L}=conv3_{\bm{\mu}}(\mathbf{L}_k),
\end{equation}
where $conca(\bigcdot)$ is concate. $DWT(\bigcdot)$ and $IDWT(\bigcdot)$ means discrete wavelet transform and its inverse. $conv3(\bigcdot)$ is a $3\times3$ convolution. 
In the preliminary refinement, the wavelet transform decomposes the reflectance component into frequency bands, which are then processed by a VSS module for restoration within each band. The reconstructed reflectance $\mathbf{R}'_k$, along with its illumination-aware modulation, is then passed through another VSS for enhanced texture alignment and restoration.

To further improve noise suppression and texture enhancement, we integrate the FVSS module into a high-order ordinary differential equation (ODE) framework, specifically a second-order Runge-Kutta (RK2) method. Compared with the traditional residual network structure—essentially a first-order Euler discretization of an ODE with non-negligible truncation error~\cite{weinan2017proposal}—the RK2 framework provides more accurate numerical solutions, which better accommodate the fine-grained requirements of noise removal and detail enhancement. To increase flexibility, we incorporate a learnable weighted gating mechanism $g_w$ into the RK2 solver. The final refined reflectance $\mathbf{R}_k$ is computed as: 
\begin{equation} \label{Eq:IGRE5}
 \mathbf{R}_k={\mathcal{R}}(\hat{\mathbf{R}}_{k}, \mathbf{L}_{k})=\hat{\mathbf{R}}_{k}+ g_w\widetilde{\mathbf{R}}_{k}+(1-g_w)\widetilde{\mathbf{R}}^1_{k}, 
\end{equation}
\begin{equation} \label{Eq:IGRE6}
g_w=S(\bm{\sigma}_g conv3(\widetilde{\mathbf{R}}_{k},\widetilde{\mathbf{R}}^1_{k})+\bm{\mu}_g), \ \widetilde{\mathbf{R}}_{k}=FVSS(\hat{\mathbf{R}}_k,{\mathbf{L}}_k), \  \widetilde{\mathbf{R}}^1_{k}=FVSS(\hat{\mathbf{R}}_k + \widetilde{\mathbf{R}}_{k},{\mathbf{L}}_k), 
\end{equation}
where $S(\bigcdot)$ is the softmax operator. $\bm{\sigma}_g$ and $\bm{\mu}_g$ are two learnable parameters in $g_w$. 
As the stages progress, UnfoldIR jointly promotes illumination smoothness, noise removal, and texture enhancement, while mitigating the intrinsic conflict between the recovery of illumination and reflectance.

\noindent\textbf{ISIC Loss}. Leveraging the multi-stage nature of DUNs, we design an inter-stage information consistency (ISIC) loss, $L_{ISIC}$, to enhance stability during the final stages of restoration. This loss ensures that small variations in illumination do not compromise essential reflectance details in the restored image, and vice versa. Hence, $L_{ISIC}$ contributes to reducing color distortion in the illumination map and preserving structural details in the reflectance component. $L_{ISIC}$ is defined as:
\begin{equation} \label{Eq:loss}
L_{ISIC}=\|\mathbf{R}_k\odot\mathbf{L}_{k-1}-\mathbf{R}_k\odot\mathbf{L}_k\|_2 + \|\nabla(\mathbf{R}_{k-1}\odot\mathbf{L}_k) - \nabla(\mathbf{R}_{k}\odot\mathbf{L}_k)\|_1. 
\end{equation}
\cref{Eq:loss} encourages consistency in the restored image rather than in the individual Retinex components, relaxing the constraint and allowing small changes that do not degrade essential information. Notably, $L_{{ISIC}}$ is a DUN-specific constraint, which enables broader applicability of UnfoldIR in the unsupervised settings. To ensure the model remains responsive to stage-specific dynamics, we apply $L_{{ISIC}}$ only during the final two stages.
Except $L_{{ISIC}}$, all other loss functions follow Uretinex~\cite{wu2022uretinex}. 

\begin{table*}[t]
\centering
\setlength{\abovecaptionskip}{0cm}
\caption{Results on the LLIE task. The best two results are in {\color[HTML]{FF0000}\textbf{red}} and {\color[HTML]{00B0F0}\textbf{blue}} fonts, respectively.}\label{table:low-light}
\resizebox{\columnwidth}{!}{
\setlength{\tabcolsep}{1.1mm}
\begin{tabular}{l|c|cc|cccc|cccc|cccc}
\toprule
                          &           & \multicolumn{2}{c|}{Efficiency }           & \multicolumn{4}{c|}{\textit{LOL-v1}}& \multicolumn{4}{c|}{\textit{LOL-v2-real}}& \multicolumn{4}{c}{\textit{LOL-v2-synthetic}}\\ \cline{3-16}
\multirow{-2}{*}{Methods} & \multirow{-2}{*}{Sources} & \cellcolor{gray!40} Para.~$\downarrow$ & \cellcolor{gray!40}FLOPs~$\downarrow$&
\cellcolor{gray!40}PSNR~$\uparrow$  & \cellcolor{gray!40}SSIM~$\uparrow$  & \cellcolor{gray!40}FID~$\downarrow$ & \cellcolor{gray!40}BIQE~$\downarrow$ & \cellcolor{gray!40}PSNR~$\uparrow$  & \cellcolor{gray!40}SSIM~$\uparrow$  & \cellcolor{gray!40}FID~$\downarrow$ & \cellcolor{gray!40}BIQE~$\downarrow$ & \cellcolor{gray!40}PSNR~$\uparrow$  & \cellcolor{gray!40}SSIM~$\uparrow$  & \cellcolor{gray!40}FID~$\downarrow$ & \cellcolor{gray!40}BIQE~$\downarrow$            \\ \midrule
URetinex~\citep{wu2022uretinex}             & CVPR22     & 0.36& 233.09          & 21.33                                 & 0.835                                 & 85.59                                 & 30.37                                 & 20.44                                 & 0.806                                 & 76.74                                 & 28.85                                 & 24.73                                 & 0.897                                 & 33.25                                 & 33.46                                 \\
UFormer~\citep{wang2022uformer}   & CVPR22    &  5.20  &   10.68   & 16.36                                 & 0.771                                 & 166.69                                & 41.06                                 & 18.82                                 & 0.771                                 & 164.41                                & 40.36                                 & 19.66                                 & 0.871                                 & 58.69                                 & 39.75                                \\
Restormer~\citep{zamir2022restormer}                 & CVPR22   &26.13 & 144.25        & 22.43                                 & 0.823                                 & 78.75                                 & 33.18                                 & 19.94                                 & 0.827                                 & 114.35                                & 37.27                                 & 21.41                                 & 0.830                                 & 46.89                                 & 35.06                                 \\
SNR-Net~\citep{xu2022snr}                   & CVPR22     &4.01 &26.35            & 24.61                                 & 0.842                                 & 66.47                                 & 28.73                                 & 21.48                                 & 0.849                                 & 68.56                                 & 28.83                                 & 24.14                                 & 0.928                                 & 30.52                                 & 33.47                                 \\
SMG~\citep{xu2023low}                       & CVPR23     & 14.02 &  17.55           & 24.82                                 & 0.838                                 & 69.47                                 & 30.15                                 & 22.62                                 & 0.857                                 & 71.76                                 & 30.32                                 & 25.62                                 & 0.905                                 & 23.36                                 & 29.35                                 \\
Diff-Retinex~\citep{yi2023diff}              & ICCV23     & 56.88 & 198.16&    21.98                                 & 0.852                                 & 51.33 & { {19.62}} & 20.17                                 & 0.826                                 & { {46.67}} & { {24.18}} & 24.30                                 & 0.921                                 & 28.74                                 & 26.35                                \\
MRQ~\citep{liu2023low}     & ICCV23    &  8.45 &  20.66      & {\color[HTML]{00B0F0} \textbf{25.24}} & 0.855 & 53.32                                 & 22.73                                 & 22.37                                 & 0.854                                 & 68.89                                 & 33.61                                 & 25.54                                 & 0.940                                 & 20.86 & 25.09 \\
IAGC~\citep{wang2023low}                      & ICCV23     &--- &---           & 24.53                                 & 0.842                                 & 59.73                                 & 25.50                                 & 22.20                                 & { {0.863}} & 70.34                                 & 31.70                                 & 25.58                                 &0.941 & 21.38                                 & 30.32 \\
DiffIR~\citep{xia2023diffir}                    & ICCV23    & 27.80& 35.32            & 23.15                                 & 0.828                                 & 70.13                                 & 26.38                                 & 21.15                                 & 0.816                                 & 72.33                                 & 29.15                                 & 24.76                                 & 0.921                                 & 28.87                                 & 27.74                                \\
CUE~\citep{zheng2023empowering}                       & ICCV23     & 0.25& 157.32   & 21.86                                 & 0.841                                 & 69.83                                 & 27.15                                 & 21.19                                 & 0.829                                 & 67.05                                 & 28.83                                 & 24.41                                 & 0.917                                 & 31.33                                 & 33.83                                \\
GSAD~\citep{jinhui2023global} & NIPS23  & 17.17& 670.33 & 23.23  &0.852  &51.64 &19.96  & 20.19 & 0.847 &46.77 &28.85 & 24.22 &0.927 &{{19.24}} &25.76 \\
AST~\citep{Zhou_2024_CVPR}  & CVPR24  & 19.90&13.25 & 21.09 &0.858 &87.67 &21.23  &21.68 &0.856 &91.81 &25.17 &22.25 &0.927 &37.19 &28.78 \\
RetiMamba~\cite{bai2024retinexmamba} &ArXiv & 3.59 & 37.98 & 24.03 & 0.827 & 75.33 & 16.28 & 22.45 & 0.844 & 56.96 & 21.76 & 25.89 & 0.934 & 20.17 & 16.29 \\
MambaIR~\citep{guo2024mambair} &ECCV24  &4.30 &60.66 & 22.23 &{{0.863}} &63.39 &20.17  &21.15 &0.857 &56.09 &24.46 &{{25.75}} &{0.937} &19.75 &{{20.37}}\\
Mamballie~\cite{weng2024mamballie} & NIPS24  &2.28 &20.85 & 23.24 & 0.861 & --- & --- & 22.95 & 0.847 & --- & --- & 25.87 & 0.940 & --- & ---   \\
Reti-Diff~\cite{he2025reti} & ICLR25     & 26.11& 87.63& {\color[HTML]{FF0000} \textbf{25.35}} & { {0.866}} & { {49.14}} & { {17.75}} & { {22.97}} & { {0.858}} & {\color[HTML]{00B0F0} \textbf{43.18}} & { {23.66}} & {\color[HTML]{00B0F0} \textbf{27.53}} & { {0.951}} & {\color[HTML]{FF0000} \textbf{13.26}} & { {15.77}} \\
CIDNet~\cite{yan2024you} & CVPR25 & 1.88& 7.57&  23.50 & {\color[HTML]{00B0F0} \textbf{0.900}} & {\color[HTML]{00B0F0} \textbf{46.69}} & {\color[HTML]{00B0F0} \textbf{14.77}} & {\color[HTML]{FF0000} \textbf{24.11}} & 0.871 & 48.04 & 18.45 & 25.71 & 0.942 & 18.60 & 15.87\\
\rowcolor{gray!10} UnfoldIR-t & Ours & 0.09 & 0.86 & 21.08 & 0.858 & 66.82 & 28.73 & 20.73 & 0.836 & 66.04 & 24.19 & 24.49 & 0.920 & 29.11 & 28.83  \\
\rowcolor{gray!10} UnfoldIR-s & Ours &0.35 &2.00 & 22.57 & 0.897 &50.37 & 15.50 & 21.64 & {\color[HTML]{00B0F0} \textbf{0.876}} & 44.65 & {\color[HTML]{00B0F0} \textbf{18.12}} & 24.92 & {\color[HTML]{00B0F0} \textbf{0.952}} & 20.05 & {\color[HTML]{00B0F0} \textbf{15.66}}   \\
\rowcolor{gray!10} UnfoldIR & Ours & 3.45&11.83 & 24.41 & {\color[HTML]{FF0000} \textbf{0.911}} & {\color[HTML]{FF0000} \textbf{43.08}} & {\color[HTML]{FF0000} \textbf{13.21}} & {\color[HTML]{00B0F0} \textbf{22.99}} & {\color[HTML]{FF0000} \textbf{0.887}} & {\color[HTML]{FF0000} \textbf{34.86}} & {\color[HTML]{FF0000} \textbf{17.84}} & {\color[HTML]{FF0000} \textbf{27.55}} & {\color[HTML]{FF0000} \textbf{0.959}} & {\color[HTML]{00B0F0} \textbf{18.07}} & {\color[HTML]{FF0000} \textbf{15.08}}  \\ \bottomrule
\end{tabular}}
\vspace{-4.5mm}
\end{table*}
\begin{figure*}[t]
\setlength{\abovecaptionskip}{0cm} 	\setlength{\belowcaptionskip}{0cm}
\includegraphics[width=1\linewidth]{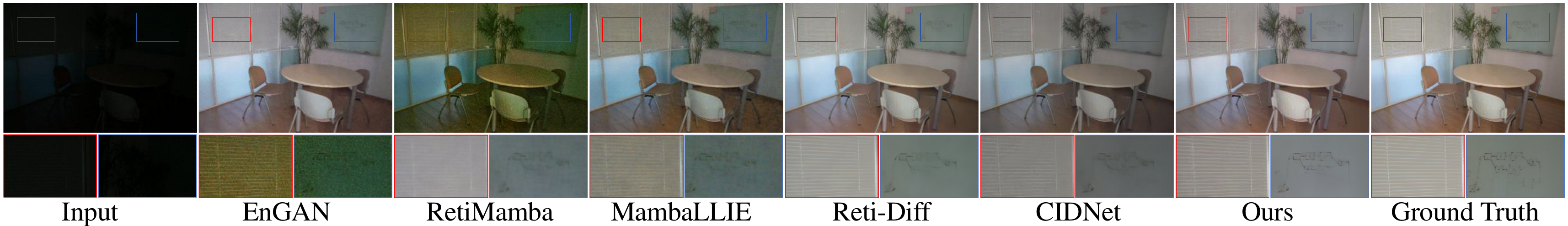}
	\captionof{figure}{Visual results on the low-light image enhancement task.}
    \label{fig:low-light}
	\vspace{-8mm}
\end{figure*}

\section{Experiments} \label{experiments}
\noindent\textbf{Experimental setup}. Our UnfoldIR is implemented in PyTorch on RTX4090 GPUs and is optimized by Adam with momentum terms (0.9, 0.999). 
Random rotation and flips are used for augmentation. The stage number $K$ is set as 3.
Other parameters inherited from traditional methods are optimized in a learnable manner. For efficiency, different stages in UnfoldIR share the same weights. For fairness,
We abandon GT-mean and all compared results are obtained using the official codes. 

\begin{table*}[t]
\begin{minipage}[c]{0.338\textwidth}
\centering
\setlength{\abovecaptionskip}{0cm}
\setlength{\belowcaptionskip}{0.05cm}
\caption{Results on the UIE task.}
\resizebox{\columnwidth}{!}{
\setlength{\tabcolsep}{0.6mm}
\begin{tabular}{l|c|cccc}
\toprule
 && \multicolumn{4}{c}{\textit{UIEB}}\\ \cline{3-6}
\multirow{-2}{*}{Methods} &\multirow{-2}{*}{Sources} & \cellcolor{gray!40}PSNR~$\uparrow$ & \cellcolor{gray!40}SSIM~$\uparrow$& \cellcolor{gray!40}UCIQE~$\uparrow$& \cellcolor{gray!40}UIQM~$\uparrow$\\ \midrule
S-uwnet~\citep{naik2021shallow}& AAAI21&18.28                                 & 0.855                                 & 0.544                                 & 2.942                                 \\
PUIE~\citep{fu2022uncertainty}& ECCV22 & 21.38                                & 0.882                                 & 0.566                                 & { {3.021}} \\
U-shape~\citep{peng2023u}& TIP23 & 22.91                                 &0.905 & 0.592                                 & 2.896                                \\
PUGAN~\citep{cong2023pugan}& TIP23& { {23.05}} & 0.897                                 & 0.608                                 & 2.902                                          \\
ADP~\citep{zhou2023underwater}& IJCV23   & 22.90                                 & 0.892                                 &0.621 & 3.005                                  \\
NU2Net~\citep{guo2023underwater}   & AAAI23 & 22.38                                 & 0.903                                 & 0.587                                 & 2.936                                 \\
AST~\citep{Zhou_2024_CVPR}  & CVPR24 & 22.19 & 0.908 & 0.602 & 2.981 \\
MambaIR~\citep{guo2024mambair} &ECCV24 &22.60 &{ {0.916}} &{{0.617}} &2.991 \\
Reti-Diff~\cite{he2025reti}              & ICLR25        & {\color[HTML]{00B0F0} \textbf{24.12}} & {\color[HTML]{00B0F0} \textbf{0.910}} & {\color[HTML]{00B0F0} \textbf{0.631}} & {\color[HTML]{00B0F0} \textbf{3.088}} \\ 
\rowcolor{gray!10} UnfoldIR & Ours & {\color[HTML]{FF0000} \textbf{24.16}} & {\color[HTML]{FF0000} \textbf{0.930}} & {\color[HTML]{FF0000} \textbf{0.651}} & {\color[HTML]{FF0000} \textbf{3.248}} \\ \bottomrule
\end{tabular}}
\label{table:Underwater} \vspace{-4.5mm}
\end{minipage}
\begin{minipage}[c]{0.328\textwidth}
\centering
\setlength{\abovecaptionskip}{0cm}
\setlength{\belowcaptionskip}{0.05cm}
\caption{Results on the BIE task.}
\resizebox{\columnwidth}{!}{
\setlength{\tabcolsep}{0.6mm}
\begin{tabular}{l|c|cccc}
\toprule
&  & \multicolumn{4}{c}{\textit{BAID}}                                                                                                                             \\ \cline{3-6}
\multirow{-2}{*}{Methods} & \multirow{-2}{*}{Sources} & \cellcolor{gray!40}PSNR~$\uparrow$ & \cellcolor{gray!40}SSIM~$\uparrow$ & \cellcolor{gray!40}LPIPS~$\downarrow$ & \cellcolor{gray!40}FID~$\downarrow$ \\ \midrule
EnGAN~\citep{jiang2021enlightengan}   & TIP21                       & 17.96                                 & 0.819                                 & 0.182                                 & 43.55                                 \\
URetinex~\citep{wu2022uretinex}    & CVPR22                      & 19.08                                 & 0.845                                 & 0.206                                 & 42.26                                 \\
CLIP-LIT~\citep{liang2023iterative}                  & ICCV23                      & 21.13                                 & 0.853                                 &0.159 &37.30 \\
Diff-Retinex~\cite{yi2023diff}              & ICCV23                      &22.07 & 0.861                                 & 0.160                                 & 38.07                                 \\
DiffIR~\cite{xia2023diffir}                    & ICCV23                      & 21.10                                 & 0.835                                 & 0.175                                 & 40.35                                 \\
AST~\cite{Zhou_2024_CVPR}  & CVPR24 &22.61 & 0.851 & 0.156 & 32.47 \\
MambaIR~\cite{guo2024mambair} &ECCV24 &{ {23.07}} &{ {0.874}} &{ {0.153}} &{ {29.13}} \\
RAVE~\cite{gaintseva2024rave} & ECCV24 & 21.26 & 0.872 & \color[HTML]{00B0F0} \textbf{0.096} & 64.89 \\
Reti-Diff~\cite{he2025reti}  & ICLR25                    & {\color[HTML]{00B0F0} \textbf{23.19}} & {\color[HTML]{00B0F0} \textbf{0.876}} & {{0.147}} & {\color[HTML]{FF0000} \textbf{27.47}} \\
\rowcolor{gray!10} UnfoldIR & Ours & {\color[HTML]{FF0000} \textbf{24.83}}  &\color[HTML]{FF0000} \textbf{0.890} &\color[HTML]{FF0000} \textbf{0.091} & {\color[HTML]{00B0F0} \textbf{34.64}} \\ \bottomrule
\end{tabular}}
\label{table:backlit} \vspace{-4.5mm}
\end{minipage}
\begin{minipage}[c]{0.314\textwidth}
\centering 
\setlength{\abovecaptionskip}{0cm}
\setlength{\belowcaptionskip}{0.05cm}
 \caption{Results on the FIE task.}
\resizebox{\columnwidth}{!}
{ \setlength{\tabcolsep}{0.6mm} 
\begin{tabular}{l|c|ccc} 
\toprule 
                          &                             & \multicolumn{3}{c}{\textit{Fundus}}                                                                                    \\  
\multirow{-2}{*}{Methods} & \multirow{-2}{*}{Resources} & \cellcolor{gray!40}BIQE~$\downarrow$ & \cellcolor{gray!40}CLIPIQA~$\uparrow$                               & \cellcolor{gray!40}FID~$\downarrow$                 \\ \midrule
SNR-Net~\cite{xu2022snr}                   & CVPR22                      & \color[HTML]{FF0000} \textbf{6.144}                                 & {0.557}                                 & 79.284                                 \\
URetinex~\citep{wu2022uretinex}               & CVPR22                      & 12.158                                & \color[HTML]{00B0F0} \textbf{0.561}                                 & 33.347                                 \\
SCI~\cite{ma2022toward}                       & CVPR22                      & 23.527                                & 0.552                                 & 85.175                                 \\
MIRNetV2~\cite{zamir2022learning}                  & TPAMI22                     & 14.925                                & 0.527                                 & 47.607                                 \\
FourLLE~\cite{wang2023fourllie} & MM23 & 7.741  & 0.508   & 28.736   \\
CUE~\cite{zheng2023empowering}                       & ICCV23                      & 11.721                                & 0.448                                 & 111.336                                 \\
NeRCO~\cite{yang2023implicit}  & ICCV23  &   17.256  &    0.451  &  95.241   \\
Reti-Diff~\cite{he2025reti}                 & ICLR25                      & 10.788                                & 0.525                                 & \color[HTML]{00B0F0} \textbf{27.637}                                 \\
CIDNet~\cite{yan2024you}                    & CVPR25                      & 10.663                                & 0.529                                 & 41.089                                 \\
\rowcolor{gray!10} UnfoldIR                      & Ours                    & {\color[HTML]{00B0F0} \textbf{6.719}} & {\color[HTML]{FF0000} \textbf{0.572}} & {\color[HTML]{FF0000} \textbf{27.398}}  \\ \bottomrule 
 \end{tabular}} 
\vspace{-4.5mm} 
 \label{table:fundus}
\end{minipage}
\end{table*}
\begin{figure*}[t]
\begin{minipage}[c]{1\textwidth}
\setlength{\abovecaptionskip}{0cm}
\setlength{\belowcaptionskip}{0.05cm}
\includegraphics[width=1\linewidth]{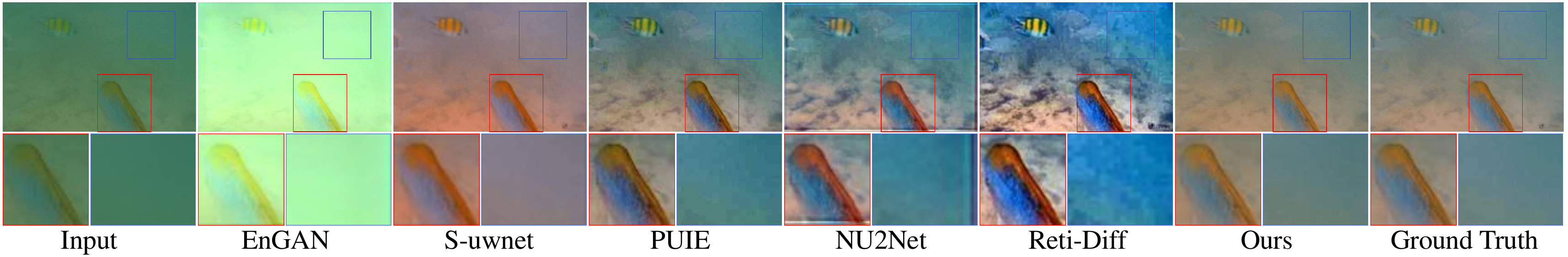}
	\caption{Visual results on the underwater image enhancement task.}
    \label{fig:underwater}
\end{minipage} \\
\begin{minipage}[c]{1\textwidth}
\setlength{\abovecaptionskip}{0cm}
\setlength{\belowcaptionskip}{0.05cm}
\includegraphics[width=1\linewidth]{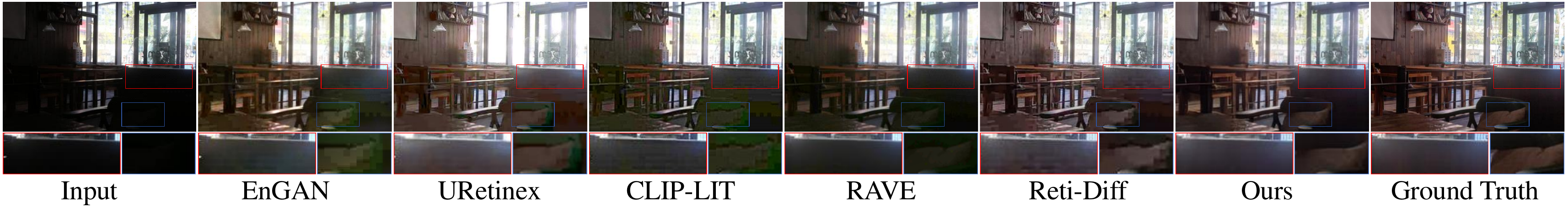}
	\caption{Visual results on the backlit image enhancement task.}
    \label{fig:backlit}
\end{minipage} \\
\begin{minipage}[c]{1\textwidth}
\setlength{\abovecaptionskip}{0cm}
\setlength{\belowcaptionskip}{0.05cm}
\includegraphics[width=1\linewidth]{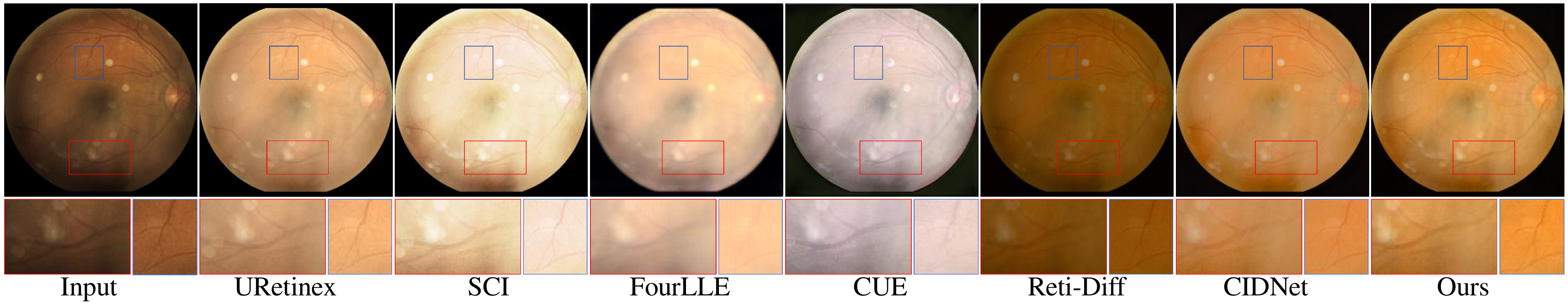}   
	\caption{Visual results on the fundus image enhancement task.}
    \label{fig:fundus}
	\vspace{-6mm}
    \end{minipage}
\end{figure*}
\subsection{Compative Evaluation}
\noindent\textbf{Low-light image enhancement}. Following Reti-Diff~\cite{he2025reti}, we report results on \textit{LOL-v1} \cite{wei2018deep}, \textit{LOL-v2-real} \cite{yang2021sparse}, and \textit{LOL-v2-syn} \cite{yang2021sparse} with four metrics: PSNR, SSIM, FID \cite{heusel2017gans}, and BIQE \cite{moorthy2010two}. Higher PSNR and SSIM values, and lower FID and BIQE scores, indicate better performance. we compare UnfoldIR with state-of-the-art techniques on $256\times256$ resolution inputs. The quantitative results are presented in Table~\ref{table:low-light}, where our method achieves top performance across all datasets while maintaining competitive efficiency.
Besides, we introduce two lightweight variants, UnfoldIR-t and UnfoldIR-s, which outperform existing lightweight models, highlighting the flexibility and scalability of UnfoldIR.
Visualizations are shown in~\cref{fig:low-light}, where our method demonstrates superiority in producing visually coherent restorations with accurately corrected illumination and enhanced textures.

\noindent\textbf{Underwater image enhancement}. We evaluate our method on \textit{UIEB}~\citep{li2019underwater}
with two metrics, UCIQE~\citep{yang2015underwater} and UIQM~\citep{panetta2015human}, where higher values indicate better results. The quantitative results are shown in Table~\ref{table:Underwater}, where our method achieves a leading place. Visualizlations in~\cref{fig:underwater} verify our effectiveness in correcting color aberrations and enhancing fine texture in underwater scenes.

\noindent\textbf{Backlit image enhancement}. Following CLIP-LIT~\citep{liang2023iterative}, we train our network using \textit{BAID} dataset~\citep{lv2022backlitnet} and evaluate with PSNR, SSIM, LPIPS~\citep{zhang2018unreasonable}, and FID~\citep{heusel2017gans}.
The results in Table~\ref{table:backlit} show that our method consistently outperforms existing approaches. Furthermore, visualizations in~\cref{fig:backlit} demonstrate our effectiveness in detail reconstruction and color correction under challenging backlit conditions.

\noindent\textbf{Fundus image enhancement}. Same as Reti-Diff~\cite{he2025reti}, we test performance on the \textit{Fundus} dataset by employing models pretrained on \textit{LOL-v2-syn} and evaluate performance on BIQE, CLIPIQA~\cite{wang2023exploring}, and FID. A larger CLIPIQA indicates a better performance. As depicted in Table~\ref{table:fundus} and~\cref{fig:fundus}, our proposed UnfoldIR consistently outperforms existing methods, both qualitatively and quantitatively.

\noindent\textbf{Real-world illumination degradation image restoration}. 
We select four real-world datasets: \textit{DICM}~\cite{lee2013contrast}, \textit{LIME}~\cite{guo2016lime}, \textit{MEF}~\cite{wang2013naturalness}, 
and \textit{VV}~\cite{he2024diffusion}. Following~\citep{feng2024you}, we employ a model pretrained on \textit{LOL-v2-syn} for inference and evaluate with PI~\cite{blau20182018} and NIQE~\cite{mittal2012making}, where lower values indicate better results. As shown in Table~\ref{table:realIDIR}, our method achieves a leading place across all datasets.

\begin{table*}[t]
\centering
\begin{minipage}[c]{0.58\textwidth}
\centering
\setlength{\abovecaptionskip}{0.05cm}
\caption{Results on the real-world IDIR task.}
\label{table:realIDIR}
\resizebox{\columnwidth}{!}{
\setlength{\tabcolsep}{1mm}
\begin{tabular}{l|c|cc|cc|cc|cc}
\toprule
\multirow{2}{*}{Methods}                            & \multicolumn{1}{c|}{\multirow{2}{*}{Sources}} & \multicolumn{2}{c|}{\textit{DICM}} & \multicolumn{2}{c|}{\textit{LIME}} & \multicolumn{2}{c|}{\textit{MEF}} & \multicolumn{2}{c}{\textit{VV}} \\
& \multicolumn{1}{c|}{}                         &\cellcolor{gray!40}PI~$\downarrow$       & \cellcolor{gray!40}NIQE~$\downarrow$              & \cellcolor{gray!40}PI~$\downarrow$      & \cellcolor{gray!40}NIQE~$\downarrow$              & \cellcolor{gray!40}PI~$\downarrow$      & \cellcolor{gray!40}NIQE~$\downarrow$              & \cellcolor{gray!40}PI~$\downarrow$      & \cellcolor{gray!40}NIQE~$\downarrow$             \\ \midrule
EnGAN~\citep{jiang2021enlightengan}                                               & TIP21                                        &   4.173    & 4.064             &   3.669     & 4.593             &     4.015     & 4.705             &   3.386  & 4.047            \\
KinD++~\citep{zhang2021beyond}                                              & IJCV21                        &        3.835     & 3.898             &        3.785       & 4.908             &       4.016       & 4.557             &       3.773       & 3.822            \\
SNR-Net~\citep{xu2022snr}                                             & CVPR22                                       &   3.585   & 4.715             &  3.753     & 5.937             &   3.677   & 6.449              & 3.503     & 9.506            \\
DCC-Net~\citep{zhang2022deep}                                             & CVPR22                                       &      3.630         & 3.709             &  {3.312}          & {4.425}             & {3.424}           & 4.598           &       3.615       & {3.286}            \\
UHDFor~\citep{UHDFourICLR2023} &ICLR23& 3.684 &4.575 & 4.124 & 4.430 &  3.813 & 4.231 & 3.319  &4.330 \\
PairLIE~\citep{fu2023learning} & CVPR23                                       & 3.685 & 4.034             &  3.387   & 4.587             &  4.133  & 4.065        & {3.334}  & 3.574            \\
GDP~\citep{fei2023generative} & CVPR23 & {3.552}  & 4.358&   4.115 & 4.891 &  3.694  & 4.609 & 3.431    & 4.683 \\
Reti-Diff~\cite{he2025reti}  & ICLR25 &\color[HTML]{00B0F0} \textbf{2.976}     &\color[HTML]{00B0F0} \textbf{3.523}    & \color[HTML]{00B0F0} \textbf{3.111}     & \color[HTML]{00B0F0} \textbf{4.128}    & {2.876} &\color[HTML]{00B0F0} \textbf{3.554}   & {2.651}    &\color[HTML]{00B0F0} \textbf{2.540}  \\ 
CIDNet~\cite{yan2024you} & CVPR25 & 3.045 & 3.796 & 3.146 & 4.132 & \color[HTML]{FF0000} \textbf{2.683} & 3.568 & \color[HTML]{00B0F0} \textbf{2.826} & 3.218 \\
\rowcolor{gray!10} UnfoldIR & Ours & \color[HTML]{FF0000} \textbf{2.952} & \color[HTML]{FF0000} \textbf{3.381} & \color[HTML]{FF0000} \textbf{3.085} & \color[HTML]{FF0000} \textbf{4.099} & \color[HTML]{00B0F0} \textbf{2.722} & \color[HTML]{FF0000} \textbf{3.387} & \color[HTML]{FF0000} \textbf{2.553} & \color[HTML]{FF0000} \textbf{2.306} \\ \bottomrule
\end{tabular}}
\vspace{-5.5mm}
\end{minipage}
\end{table*}

\begin{table}[t]
\centering
\begin{minipage}[c]{1\textwidth}
\caption{ Ablation study in the LLIE task.}
		\label{table:ablation}
\setlength{\abovecaptionskip}{0.05cm}
\resizebox{\columnwidth}{!}{
\setlength{\tabcolsep}{0.8mm}
\begin{tabular}{l|c|ccc|cccc|ccc|c}
\toprule
\multirow{2}{*}{Datasets}      & \multirow{2}{*}{Metrics} & \multicolumn{3}{c|}{Effect of RAIC}                                          & \multicolumn{4}{c|}{Effect of IGRE}                                                                   & \multicolumn{3}{c|}{Effect of $L_{ISIC}$}  & UnfoldIR       \\ \cline{3-12}
                               &                          & \cellcolor{c2!50}$\mathcal{L}_1(\bigcdot) \rightarrow \mathcal{L}(\bigcdot)$ & \cellcolor{c2!50}$\mathcal{L}_2(\bigcdot) \rightarrow \mathcal{L}(\bigcdot)$ & \cellcolor{c2!50}$\mathcal{L}_3(\bigcdot) \rightarrow \mathcal{L}(\bigcdot)$ & \cellcolor{c2!50}w/o RK2 & \cellcolor{c2!50}{w/o $g_w$} & \cellcolor{c2!50}w/o $L_k$ &\cellcolor{c2!50} VSS $\rightarrow$ FVSS & \cellcolor{c2!50}{w/o $L_{ISIC}$} & \cellcolor{c2!50}{$L_{ISIC}^1$} & \cellcolor{c2!50}{$ L_{ISIC}^2$} & (Ours)         \\ \midrule
\multirow{2}{*}{\textit{L-v2-s}} & PSNR~$\uparrow$                      & 26.85                   & 27.51                   & \textbf{27.58}          & 27.31     & 27.23                         & 26.93     & 26.74                  & 27.37                              & 27.52                                                    & 27.50                                                    & 27.55          \\
& SSIM~$\uparrow$                      & 0.943                   & 0.956                   & 0.958                   & 0.954    & 0.955                         & 0.946     & 0.942                  & 0.943                              & 0.954                                                    & 0.952                                                    & \textbf{0.959} \\ \midrule
\multirow{2}{*}{\textit{L-v2-r}} & PSNR~$\uparrow$                      & 22.11                   & 22.87                   & 22.96                   & 22.78      & 22.87                         & 22.06     & 22.06                  & 22.26                              & 22.82                                                    & 22.88                                                    & \textbf{22.99}          \\
& SSIM~$\uparrow$                     & 0.882                   & \textbf{0.892}          & 0.888                   & 0.883         & 0.885                         & 0.876     & 0.879                  & 0.874                              & 0.884                                                    & 0.881                                                    & 0.887   \\ \bottomrule      
\end{tabular}}
\end{minipage}\\
\begin{minipage}[c]{0.573\textwidth}
\centering
\caption{Performance of UnfoldIR with different restoration models and stage numbers.}\label{table:ablationModel}
\setlength{\abovecaptionskip}{0.05cm}
\resizebox{\columnwidth}{!}{
\setlength{\tabcolsep}{0.6mm}
\begin{tabular}{l|c|ccccc|ccc|c}
\toprule
\multirow{2}{*}{Datasets}      & \multirow{2}{*}{Metrics} & \multicolumn{5}{c|}{Restoration models}       & \multicolumn{3}{c|}{Stage numbers} & \multicolumn{1}{c}{UnfoldIR} \\ \cline{3-10}
                               &                          & \cellcolor{c2!50}CM1   &\cellcolor{c2!50} CM2   &\cellcolor{c2!50} CM3   &\cellcolor{c2!50} CM4   &\cellcolor{c2!50} CM5   &\cellcolor{c2!50} K=2        & \cellcolor{c2!50} K=4        &\cellcolor{c2!50} K=5            & IDIRM,K=3                    \\ \midrule
\multirow{2}{*}{\textit{L-v2-s}} & PSNR~$\uparrow$                     & 26.32 & 27.13 & 27.08 & 26.73 & 27.25 & 26.89      & 28.32      &   \textbf{28.50}      & 27.55                        \\
& SSIM~$\uparrow$                     & 0.938 & 0.946 & 0.944 & 0.949 & 0.952 & 0.949      & 0.968      & {\textbf{0.972}}    & 0.959    \\  \midrule
\multirow{2}{*}{\textit{L-v2-r}} & PSNR~$\uparrow$                     & 21.23 & 21.94 & 22.12 & 22.30 & 22.17 & 22.57      & 23.86      & {\textbf{24.15}}   & 22.99                        \\ 
& SSIM~$\uparrow$                     & 0.860 & 0.873 & 0.872 & 0.870 & 0.882 & 0.881      & 0.893      & {\textbf{0.896}}   & 0.887    \\ \bottomrule                  
\end{tabular}}\vspace{-6mm}
\end{minipage}
\begin{minipage}[c]{0.417\textwidth}
\centering
\caption{Experiments on the unsupervised setting.}\label{table:unsupervised}
\setlength{\abovecaptionskip}{0.05cm}
\resizebox{\columnwidth}{!}{
\setlength{\tabcolsep}{1.4mm}
\begin{tabular}{cccc}
\toprule
\cellcolor{c2!50}NeRCO~\cite{yang2023implicit} & \cellcolor{c2!50}CLIP-LIT~\cite{liang2023iterative} & \cellcolor{c2!50}LightenDiff~\cite{jiang2024lightendiffusion}        & \cellcolor{c2!50}UnfoldIR       \\  
ICCV23 & ICCV23 & ECCV24 & Ours\\ \midrule
19.14 & 16.18    & 20.44          & \textbf{20.47} \\
0.743 & 0.792    & 0.843          & \textbf{0.852} \\ \midrule
19.23 & 17.06    & \textbf{19.32} & 19.28          \\
0.671 & 0.589    & 0.684          & \textbf{0.686}  \\ \bottomrule
\end{tabular}}\vspace{-6mm}
\end{minipage}
\end{table}
\subsection{Ablation Study and Further Analysis}\label{ablation-study}

\noindent\textbf{Effect of basic network components in UnfoldIR}. We conduct experiments on Table~\ref{table:ablation} to verify the effectiveness of the core components in UnfoldIR, including RAIC, IGRE, and $L_{ISIC}$. 
We first replace the VSS module, \textit{i.e.}, $\mathcal{L}$, with three alternatives: $\mathcal{L}_1$, $\mathcal{L}_2$, and $\mathcal{L}_3$. $\mathcal{L}_1$ is a transformer block from Reti-Diff~\cite{he2025reti} with comparable parameters; $\mathcal{L}_2$ and $\mathcal{L}_3$ correspond to more advanced VSS modules taken from RetiMamba~\cite{bai2024retinexmamba} and Mamballie~\cite{weng2024mamballie}. As shown in Table~\ref{table:ablation}, VSS, compared to them, can better balance performance and efficiency. 
Then we assess the effect of individual components within IGRE, including RK2, the gated mechanism $g_w$, the illumination-aware texture enhancement mechanism, and the proposed FVSS module. 
Additionally, we evaluate the contribution of our $L_{ISIC}$ by (1) removing it entirely; (2) conducting consistency constraints directly to the reflectance and illumination components instead of the restored image ($L_{ISIC}^1$); and (3) enforcing consistency across all stages rather than only the final ones (($L_{ISIC}^2$)). These results highlight the importance of the placement and formulation of ISIC loss.

\begin{table*}[t]
\begin{minipage}[c]{1\textwidth}
\centering
\caption{ Potential applications of UnfoldIR. In (a), ``-R'' and ``+'' means refining the method with UnfoldIR and integrating UnfoldIR with the method for end-to-end training. In (b), ``Comb.'' means aggregating the outputs of each stage in UnfoldIR, while BLO is short for bi-level optimization~\cite{he2023hqg}.}
		\label{table:applications}
        \vspace{-2mm}
\begin{subtable}[c]{0.566\textwidth}
\centering
\setlength{\abovecaptionskip}{0.05cm}
\caption{ Enhancing restoration performance. }
		\label{table:VisualApplications}
\resizebox{\columnwidth}{!}{
\setlength{\tabcolsep}{0.6mm}
\begin{tabular}{l|c|ccc|ccc}
\toprule
Datasets                       & Metrics &\cellcolor{c2!50} Reti-Diff &\cellcolor{c2!50} Reti-Diff-R &\cellcolor{c2!50} Reti-Diff+ & \cellcolor{c2!50} CIDNet &\cellcolor{c2!50} CIDNet-R &\cellcolor{c2!50} CIDNet+ \\ \midrule
\multirow{2}{*}{\textit{L-v2-s}} & PSNR~$\uparrow$    & 27.53     & 27.55       & 28.82      & 25.71  & 25.86    & 26.55   \\
& SSIM~$\uparrow$    & 0.951     & 0.958       & 0.973      & 0.942  & 0.946    & 0.958   \\ \midrule
\multirow{2}{*}{\textit{L-v2-r}} & PSNR~$\uparrow$    & 22.97     & 23.08       & 23.67      & 24.11  & 24.10    & 24.37   \\
& SSIM~$\uparrow$    & 0.858     & 0.862       & 0.872      & 0.871  & 0.875    & 0.883  \\ \bottomrule
\end{tabular}}
\end{subtable}
\begin{subtable}[c]{0.428\textwidth}
\centering
\setlength{\abovecaptionskip}{0.05cm}
\caption{ Promoting downstream tasks. }
		\label{table:DownstreamApplications}
\resizebox{\columnwidth}{!}{
\setlength{\tabcolsep}{0.6mm}
\begin{tabular}{l|c|cccc}
\toprule
Tasks                & Metrics &\cellcolor{c2!50} UnfoldIR & \cellcolor{c2!50} + Comb. &\cellcolor{c2!50} + BLO  &\cellcolor{c2!50} + Comb. \& BLO \\ \midrule
Detection            & mAP~$\uparrow$     & 78.9     & 79.3  & 79.8  & 80.3      \\ \midrule
Seman. Seg.          & mIoU~$\uparrow$    & 62.5     & 63.2  & 63.3  & 64.3      \\ \midrule
Concealed & $F_\beta$~$\uparrow$       & 0.716    & 0.724 & 0.726 & 0.731     \\
Object Seg.  & $E_\phi$~$\uparrow$       & 0.875    & 0.882 & 0.882 & 0.886     \\ \bottomrule
\end{tabular}}
\end{subtable}
\end{minipage}\\
\begin{minipage}[c]{0.276\textwidth}
\centering
\setlength{\abovecaptionskip}{0.05cm}
\caption{ User study. }
		\label{table:UserStudy} 
\resizebox{\columnwidth}{!}{
\setlength{\tabcolsep}{0.6mm}
\begin{tabular}{l|cccc}
\toprule
Methods &\multicolumn{1}{c}{\cellcolor{c2!50} \textit{L-v1}}           & \multicolumn{1}{c}{\cellcolor{c2!50} \textit{L-v2}}      & \multicolumn{1}{c}{\cellcolor{c2!50} \textit{UIEB}}       & \multicolumn{1}{c}{\cellcolor{c2!50} \textit{BAID}}   \\ \midrule
Uretinex  & 3.69                                 & 3.63                                 & ---                                    & 3.13                                 \\
Restormer & 3.17                                 & 3.20                                 & ---                                    & 3.08                                 \\
SNR       & 3.73                                 & 3.88                                 & ---                                    & 3.25                                 \\
CUE       & 3.30                                 & 3.57                                 & ---                                    & ---                                    \\
AST       & 3.58                                 & 3.65                                 & 3.82                                 & 3.37                                 \\
MambaIR   & 3.70                                 & 3.77                                 & 4.03                                 & 3.45                                 \\
Reti-Diff & {\color[HTML]{00B0F0} \textbf{3.89}} & {\color[HTML]{00B0F0} \textbf{4.13}} & {\color[HTML]{00B0F0} \textbf{4.18}} & {\color[HTML]{00B0F0} \textbf{3.60}} \\
CIDNet    & 3.67                                 & 3.82                                 & ---                                   & ---                                    \\
\rowcolor{gray!10} Ours  & {\color[HTML]{FF0000} \textbf{4.12}} & {\color[HTML]{FF0000} \textbf{4.17}} & {\color[HTML]{FF0000} \textbf{4.34}} & {\color[HTML]{FF0000} \textbf{3.75}} \\ \bottomrule
\end{tabular}}
\end{minipage}
\begin{minipage}[c]{0.72\textwidth}
\centering
\setlength{\abovecaptionskip}{0.05cm}
\caption{Low-light image detection on \textit{ExDark}.
}
		\label{table:Detection} 
\resizebox{\columnwidth}{!}{
\setlength{\tabcolsep}{0.73mm}
\begin{tabular}{l|cccccccccccc|c}
\toprule
Methods (AP)& \cellcolor{c2!50}Bicycle  & \cellcolor{c2!50}Boat & \cellcolor{c2!50}Bottle  & \cellcolor{c2!50}Bus & \cellcolor{c2!50}Car & \cellcolor{c2!50}Cat  & \cellcolor{c2!50}Chair &\cellcolor{c2!50} Cup & \cellcolor{c2!50}Dog & \cellcolor{c2!50} Motor & \cellcolor{c2!50}People & \cellcolor{c2!50}Table  & \cellcolor{c2!50}Mean \\ \midrule
Baseline & 74.7                                 & 64.9                                 & 70.7                                 & 84.2                                 & 79.7                                 & 47.3                                 & 58.6                                 & 67.1                                 & 64.1                                 & 66.2                                 & 73.9                                 & 45.7                                 & 66.4                                 \\
RetinexNet
& 72.8                                 & 66.4                                 & 67.3                                 & 87.5                                 & 80.6                                 & 52.8                                 & 60.0                                 & 67.8                                 & 68.5                                 & {{69.3}} & 71.3                                 & 46.2                                 & 67.5                                 \\
KinD  & 73.2                                 & 67.1                                 & 64.6                                 & 86.8                                 & 79.5                                 & 58.7                                 & 63.4                                 & 67.5                                 & 67.4                                 & 62.3                                 & {{75.5}} & 51.4                                 & 68.1                                 \\
MIRNet & 74.9                                 & 69.7                                 & 68.3                                 & 89.7                                 & 77.6                                 & 57.8                                 & 56.9                                 & 66.4                                 & 69.7                                 & 64.6                                 & 74.6                                 & 53.4                                 & 68.6                                 \\
RUAS   & 75.7                                 & 71.2                                 & 73.5                                 & \color[HTML]{00B0F0} \textbf{90.7}                                 & 80.1                                 & 59.3                                 & {{67.0}} & 66.3                                 & 68.3                                 & 66.9                                 & 72.6                                 & 50.6                                 & 70.2                                 \\
SCI   & 73.4                                 & 68.0                                 & 69.5                                 & 86.2                                 & 74.5                                 & 63.1                                 & 59.5                                 & 61.0                                 & 67.3                                 & 63.9                                 & 73.2                                 & 47.3                                 & 67.2                                 \\
SNR-Net   & {{78.3}} & 74.2                                 & {{74.5}} & 89.6                                 & {\color[HTML]{00B0F0} \textbf{82.7}} & {{66.8}} & 66.3                                 & 62.5                                 & 74.7                                 & 63.1                                 & 73.3                                 & {{57.2}} & 71.9                                 \\
Reti-Diff & {\color[HTML]{00B0F0} \textbf{82.0}} & {\color[HTML]{00B0F0} \textbf{77.9}} & {\color[HTML]{00B0F0} \textbf{76.4}} & {\color[HTML]{FF0000} \textbf{92.2}} & {\color[HTML]{FF0000} \textbf{83.3}} & {\color[HTML]{00B0F0} \textbf{69.6}} & {\color[HTML]{00B0F0} \textbf{67.4}} & {\color[HTML]{00B0F0} \textbf{74.4}} & {\color[HTML]{00B0F0} \textbf{75.5}} & {\color[HTML]{00B0F0} \textbf{74.3}} & {\color[HTML]{00B0F0} \textbf{78.3}} & {\color[HTML]{00B0F0} \textbf{57.9}} & {\color[HTML]{00B0F0} \textbf{75.8}}\\
\rowcolor{gray!10} Ours & \multicolumn{1}{c}{{\color[HTML]{FF0000} \textbf{87.5}}} & \multicolumn{1}{c}{{\color[HTML]{FF0000} \textbf{81.0}}} & \multicolumn{1}{c}{{\color[HTML]{FF0000} \textbf{78.2}}} & \multicolumn{1}{c}{{ {86.4}}} & \multicolumn{1}{c}{{ {74.6}}} & \multicolumn{1}{c}{{\color[HTML]{FF0000} \textbf{76.4}}} & \multicolumn{1}{c}{{\color[HTML]{FF0000} \textbf{80.1}}} & \multicolumn{1}{c}{{\color[HTML]{FF0000} \textbf{80.8}}} & \multicolumn{1}{c}{{\color[HTML]{FF0000} \textbf{83.4}}} & \multicolumn{1}{c}{{\color[HTML]{FF0000} \textbf{84.1}}} & \multicolumn{1}{c}{{\color[HTML]{FF0000} \textbf{69.1}}} & \multicolumn{1}{c|}{{\color[HTML]{FF0000} \textbf{65.0}}} & \multicolumn{1}{c}{{\color[HTML]{FF0000} \textbf{78.9}}}
\\ \bottomrule
\end{tabular}}
\end{minipage}\\
\begin{minipage}[c]{0.66\textwidth}
\centering
\setlength{\abovecaptionskip}{0.05cm}
\caption{Low-light semantic segmentation, where images are darkened by~\citep{zhang2021learning}. }
		\label{table:SemanticSegmentation} 
\resizebox{\columnwidth}{!}{
\setlength{\tabcolsep}{0.8mm}
\begin{tabular}{l|cccccccccc|c}
\toprule
Methods (IoU)        & \cellcolor{c2!50}Bicycle                               & \cellcolor{c2!50}Boat                                  & \cellcolor{c2!50}Bottle                                & \cellcolor{c2!50}Bus                                   & \cellcolor{c2!50}Car                                   & \cellcolor{c2!50}Cat                                   & \cellcolor{c2!50}Chair                                 & \cellcolor{c2!50}Dog                                   & \cellcolor{c2!50}Horse                                 & \cellcolor{c2!50}People                                & \cellcolor{c2!50}Mean                                  \\ \midrule
Baseline& 43.5                                 & 36.3                                 & 48.6                                 & 70.5                                 & 67.3                                 & 46.6                                 & 11.2                                 & 42.4                                 & 56.7                                 & 57.8                                 & 48.1                                   \\
RetinexNet   & 48.6                                 & 41.7                                 & 51.7                                 & 77.6                                 & 68.3                                 & 52.7                                 & 15.8                                 & 46.3                                 & 60.2                                 & 62.3                                 & 52.5                                 \\
KinD   & 51.3                                 & 40.2                                 & 53.2                                 & 76.8                                 & 69.4                                 & 50.8                                 & 14.6                                 & 47.3                                 & 60.3                                 & 60.9                                 & 52.5                                 \\
MIRNet   & 50.3                                 & 42.9                                 & 47.4                                 & 73.6                                 & 62.7                                 & 50.4                                 & 15.8                                 & 46.3                                 & 61.0                                 & 63.3                                 & 51.4                                 \\
RUAS    & 53.0                                 & 37.3                                 & 50.4                                 & 71.3                                 & 72.3                                 & 47.6                                 & 15.9                                 & 50.8                                 & 63.6                                 & 60.8                                 & 52.3                                 \\
SCI    & 54.5                                 & 46.3                                 & 57.2                                 & 78.4                                 & 73.3                                 & 49.1                                 & 22.8                                 & 49.0                                 & 62.1                                 & 66.9                                 & 56.0                                 \\
SNR-Net   & {{57.7}} & {{48.6}} & {{59.5}} & {{81.3}} & {{74.8}} & 50.2                                 & {{24.4}} & 50.7                                 & {{64.3}} & 68.7                                 & {{58.0}} \\
Reti-Diff & {\color[HTML]{00B0F0} \textbf{59.8}} & {\color[HTML]{00B0F0} \textbf{51.5}} & {\color[HTML]{FF0000} \textbf{62.1}} & {\color[HTML]{FF0000} \textbf{85.5}} & {\color[HTML]{00B0F0} \textbf{76.6}} & {\color[HTML]{00B0F0} \textbf{57.7}} & {\color[HTML]{00B0F0} \textbf{28.9}} & {\color[HTML]{00B0F0} \textbf{56.3}} & {\color[HTML]{00B0F0} \textbf{66.2}} & {\color[HTML]{00B0F0} \textbf{73.4}} & {\color[HTML]{00B0F0} \textbf{61.8}} \\
Ours      & {\color[HTML]{FF0000} \textbf{60.2}} & {\color[HTML]{FF0000} \textbf{51.8}} & {\color[HTML]{00B0F0} \textbf{61.3}} & {\color[HTML]{00B0F0} \textbf{84.7}} & {\color[HTML]{FF0000} \textbf{78.5}} & {\color[HTML]{FF0000} \textbf{58.8}} & {\color[HTML]{FF0000} \textbf{30.2}} & {\color[HTML]{FF0000} \textbf{57.5}} & {\color[HTML]{FF0000} \textbf{66.8}} & {\color[HTML]{FF0000} \textbf{75.2}} & {\color[HTML]{FF0000} \textbf{62.5}} \\ \bottomrule
\end{tabular}}
\vspace{-6mm}
\end{minipage}
\begin{minipage}[c]{0.333\textwidth}
\centering
\setlength{\abovecaptionskip}{0.05cm}
\caption{Low-light concealed object segmentation.
}
		\label{table:CODSegmentation}
\resizebox{\columnwidth}{!}{
\setlength{\tabcolsep}{1.4mm}
\begin{tabular}{l|cccc}
\toprule
{Methods} & \cellcolor{gray!40}$M$~$\downarrow$ & \cellcolor{gray!40}$F_\beta$~$\uparrow$ & \cellcolor{gray!40}$E_\phi$~$\uparrow$ & \cellcolor{gray!40}$S_\alpha$~$\uparrow$ \\ \midrule
Baseline   & 0.049                                 & 0.631                                 & 0.818                                 & 0.762 \\
RetinexNet & 0.041                                 & 0.663                                 & 0.847                                 & 0.789   \\
KinD       & 0.038                                 & 0.670                                 & 0.855                                 & 0.793                                 \\
MIRNet     & 0.036                                 & 0.701                                 & 0.860                                 & 0.800                                 \\
RUAS       & 0.037                                 & 0.707                                 & 0.866                                 & 0.805                                 \\
SNR-Net    & 0.035                                 & {\color[HTML]{00B0F0} \textbf{0.708}} & 0.857                                 & {\color[HTML]{00B0F0} \textbf{0.807}} \\
SCI        & 0.036                                 & 0.702                                 & {\color[HTML]{00B0F0} \textbf{0.867}} & 0.802                                 \\
Reti-Diff  & {\color[HTML]{00B0F0} \textbf{0.034}} & {\color[HTML]{00B0F0} \textbf{0.708}} & {\color[HTML]{00B0F0} \textbf{0.867}} & {\color[HTML]{FF0000} \textbf{0.809}} \\
Ours       & {\color[HTML]{FF0000} \textbf{0.033}} & {\color[HTML]{FF0000} \textbf{0.716}} & {\color[HTML]{FF0000} \textbf{0.875}} & {\color[HTML]{00B0F0} \textbf{0.807}} \\ \bottomrule
\end{tabular}}
 \vspace{-6mm}
\end{minipage}
\end{table*}

\noindent\textbf{Other configurations in UnfoldIR}. We explore how the restoration model and stage number influence the performance. As shown in Table~\ref{table:ablationModel}, we first conduct breakdown ablations of our IDIRM model by removing two proposed explicit constraints (CM1), removing only the explicit restriction of $\mathcal{R}$ (CM2) or $\mathcal{L}$ (CM3). 
CM4 and CM5 correspond to the restoration model used in URetinex~\cite{wu2022uretinex} and CUE~\cite{zheng2023empowering}, further validating the effectiveness of our proposed restoration model. 
Besides, we explore the influence of stage number and discover that increasing the stage number can improve performance. We discover that UnfoldIR achieves cutting-edge performance when $K$ reaches 3, thus setting $K=3$.

\noindent\textbf{Potential applications of UnfoldIR}. 
We further explore the potential of the UnfoldIR framework, including its adaptability to different supervision paradigms, its positive impact on existing methods, and its benefits for downstream tasks.
First, as shown in Table~\ref{table:unsupervised}, we extend UnfoldIR to the unsupervised setting following LightenDiff~\cite{jiang2024lightendiffusion}. 
Notably, even when supervised solely by the proposed $L_{\text{ISIC}}$ loss, our method achieves performance comparable to existing SOTA unsupervised methods, highlighting the effect of $L_{\text{ISIC}}$ as a framework-specific loss function.
Next, as shown in Table~\ref{table:VisualApplications}, we examine how UnfoldIR can enhance existing restoration methods by (1) directly using UnfoldIR with our pretrain model as a refiner method by inputting the enhanced result of existing methods into UnfoldIR(``-R''), and (2) integrating our UnfoldIR after the existing methods and then end-to-end training the combined network(``+''). We observe performance gains from both settings. 
Finally, in Table~\ref{table:DownstreamApplications}, we explore how to better facilitate the performance of downstream tasks. ``Comb.'' refers to combining enhanced outputs from multiple UnfoldIR stages as a form of enhancement-based data augmentation, thus improving downstream tasks. When further combining this strategy with the bi-level optimization (BLO) framework, we can observe further performance gains.

\subsection{User Study and Downstream Tasks}
\noindent\textbf{User Study}. We conduct a user study to evaluate the visual quality of IDIR methods, including LLIE (\textit{L-v1} and \textit{L-v2}), UIE (\textit{UIEB}), and BIE (\textit{BAID}). In this study, 29 participants rated enhanced images on a scale of 1 (worst) to 5 (best) based on four criteria: (1) the presence of underexposed or overexposed regions; (2) the degree of color distortion; (3) the occurrence of unwanted noise or artifacts; and (4) the preservation of essential structural details. Each low-light image was displayed alongside its enhanced version with the method's name concealed. As shown in Table~\ref{table:UserStudy}, our method outperforms across all four datasets, verifying its effectiveness in producing visually pleasing results.

\noindent\textbf{Low-light Object Detection}. Enhanced images are expected to facilitate subsequent tasks. We first evaluate the impact on low-light object detection to test this hypothesis. Following the protocol in~\cite{he2025reti}, all compared methods are assessed on \textit{ExDark}~\cite{loh2019getting} using YOLO.
``Baseline'' corresponds to the original low-quality images without enhancement. As shown in Table~\ref{table:Detection}, our UnfoldIR outperforms competing methods, confirming its effectiveness in enhancing high-level vision performance.

\noindent\textbf{Low-light Image Segmentation}. We also performed segmentation tasks by retraining segmentation models for each enhancement method, following that employed for detection. \textbf{(1)} For semantic segmentation, following ~\citep{he2025reti}, we apply image darkening to samples from \textit{VOC}~\citep{everingham2010pascal} and utilize Mask2Former~\citep{cheng2022masked} to segment the enhanced images, evaluating the performance using Intersection over Union (IoU). As shown in Table~\ref{table:SemanticSegmentation}, our approach achieves a leading place across most classes. 
\textbf{(2)} Besides, we explore concealed object segmentation, a challenging task aimed at delineating objects with visual similarity to their backgrounds. This evaluation was conducted on \textit{COD10K}~\citep{fan2021concealed}. 
We similarly apply image darkening and employ RUN~\citep{he2025run} to segment the enhanced images. Performance was assessed using four metrics: mean absolute error ($M$), adaptive F-measure ($F_\beta$), mean E-measure ($E_\phi$), and structure measure ($S_\alpha$). As reported in Table~\ref{table:CODSegmentation}, our method outperforms existing methods.

\section{Conclusions}\vspace{-2mm}
In this paper, we propose a novel DUN-based method, UnfoldIR, for IDIR tasks. 
UnfoldIR introduces a new IDIR model with specifically designed regularization terms for smoothing illumination and enhancing texture. 
By unfolding into a multistage network, we get RAIC and IGRE modules in each stage. RAIC employs VSS to attract non-local features for color correction, while IGRE introduces a frequency-aware VSS to globally align similar textures, enhancing details. We also propose an ISIC loss to maintain network stability in the final stages. Abundant experiments verify our effectiveness.

\newpage
\bibliographystyle{unsrt}
\bibliography{egbib}

\newpage
\appendix

\setcounter{figure}{0}
\renewcommand{\figurename}{Fig.}
\renewcommand{\thefigure}{A\arabic{figure}}
\setcounter{table}{0}
\renewcommand{\tablename}{Table}
\renewcommand{\thetable}{A\arabic{table}}
\renewcommand{\thealgorithm}{A\arabic{algorithm}}

\begin{figure*}[t]
\begin{minipage}[c]{1\textwidth}
\setlength{\abovecaptionskip}{0cm}
\setlength{\belowcaptionskip}{0.05cm}
\includegraphics[width=1\linewidth]{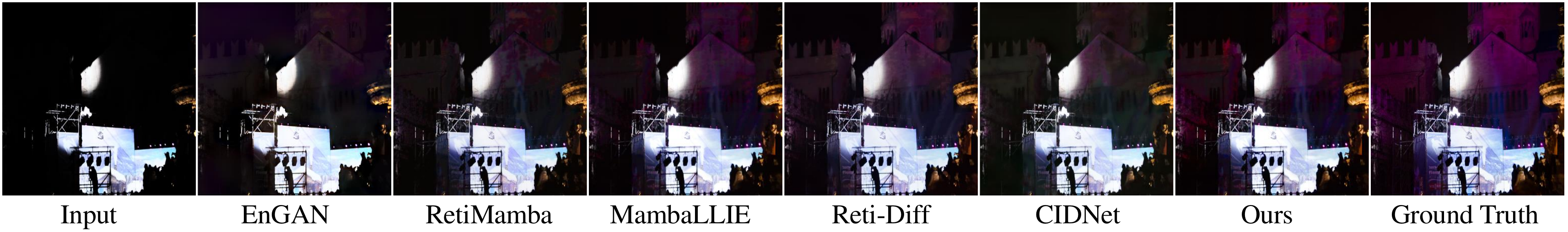}
	\caption{More visual results on the low-light image enhancement task.}
    \label{fig:llie-supp}
\end{minipage} \\
\begin{minipage}[c]{1\textwidth}
\setlength{\abovecaptionskip}{0cm}
\setlength{\belowcaptionskip}{0.05cm}
\includegraphics[width=1\linewidth]{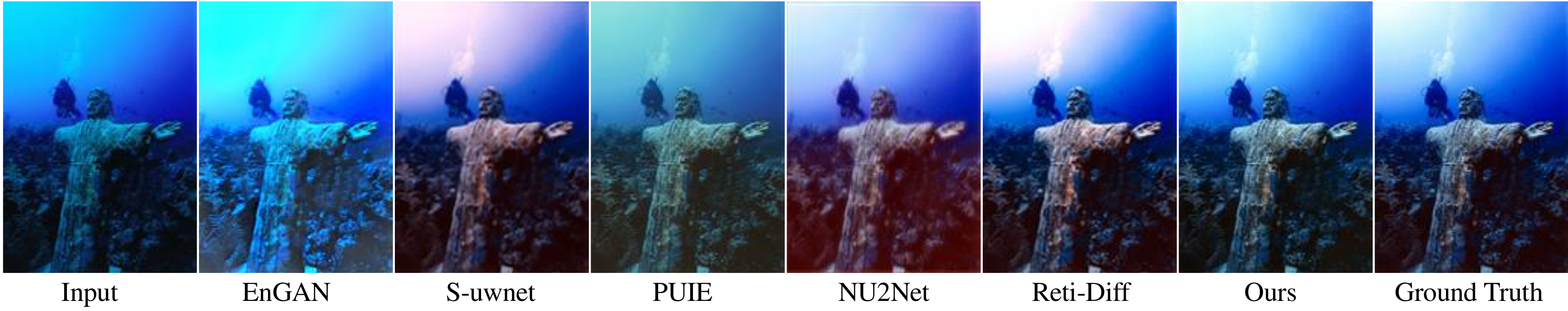}
	\caption{More visual results on the underwater image enhancement task.}
    \label{fig:underwater-supp}
\end{minipage} \\
\begin{minipage}[c]{1\textwidth}
\setlength{\abovecaptionskip}{0cm}
\setlength{\belowcaptionskip}{0.05cm}
\includegraphics[width=1\linewidth]{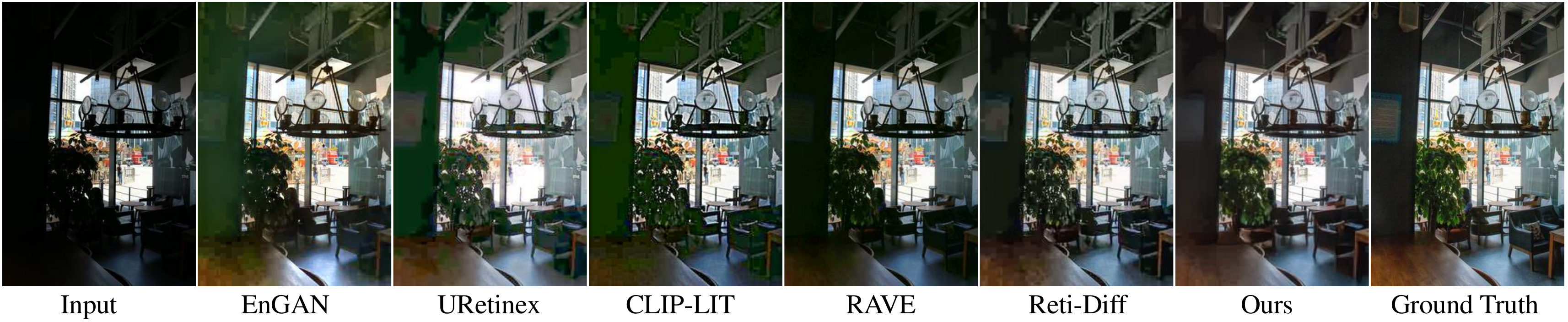}
	\caption{More visual results on the backlit image enhancement task.}
    \label{fig:backlit-supp}
\end{minipage} \\
\begin{minipage}[c]{1\textwidth}
\setlength{\abovecaptionskip}{0cm}
\setlength{\belowcaptionskip}{0.05cm}
\includegraphics[width=1\linewidth]{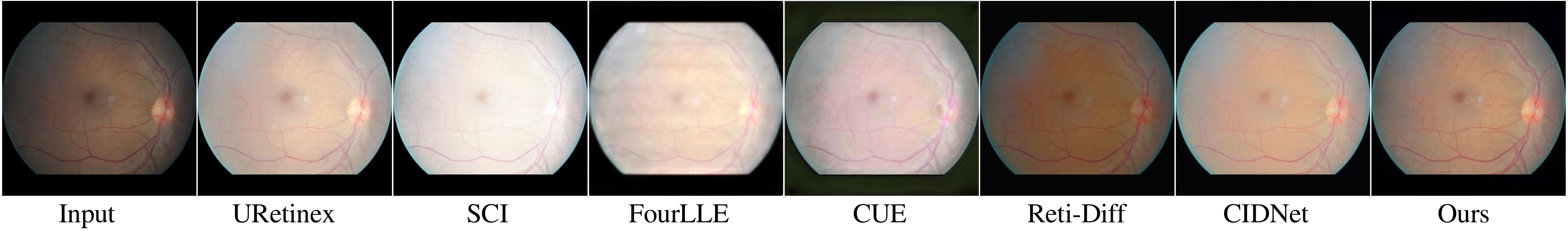}   
	\caption{More visual results on the fundus image enhancement task.}
    \label{fig:fundus-supp}
	\vspace{-6mm}
    \end{minipage}
\end{figure*}
\begin{table}[htbp!]
\centering
\caption{Exploration of the intrinsic advantages of DUNs and investigation of the deployment for image restoration. where ``U'' denotes our UnfoldIR and the suffix ``-'' indicates constructing the network based on the basic restoration model, \textit{i.e.}, CM1 in Table~\ref{table:ablationModel}. In (b), EC refers to extra connections. (c) and (d) aim to evaluate the generalizability of our $L_{ISIC}$ loss to additional tasks.}
		\label{table:discussions}
        \vspace{-2mm}
\begin{subtable}[c]{0.358\textwidth}
\centering
\setlength{\abovecaptionskip}{0.05cm}
\caption{ \fontsize{7.1pt}{\baselineskip}\selectfont Regularization terms. }
		\label{table:DiscussionRT}
\resizebox{\columnwidth}{!}{
\setlength{\tabcolsep}{0.6mm}
\begin{tabular}{l|l|cc|cc|cc}
\toprule
Datasets                       & Metrics & U-t-  & U-t   & U-s-  & U-s   & U-    & U     \\ \midrule
\multirow{2}{*}{\textit{v2-s}} & PSNR~$\uparrow$    & 20.08 & 24.49 & 22.36 & 24.92 & 26.32 & 27.55 \\
& SSIM~$\uparrow$    & 0.886 & 0.920 & 0.929 & 0.952 & 0.938 & 0.959 \\ \midrule
\multirow{2}{*}{\textit{v2-r}} & PSNR~$\uparrow$    & 18.89 & 20.73 & 20.38 & 21.64 & 21.23 & 22.99 \\
& SSIM~$\uparrow$    & 0.806 & 0.836 & 0.868 & 0.876 & 0.860  & 0.887 \\ \bottomrule
\end{tabular}}
\end{subtable}
\begin{subtable}[c]{0.165\textwidth}
\centering
\setlength{\abovecaptionskip}{0.05cm}
\caption{ \fontsize{7.1pt}{\baselineskip}\selectfont Extra connections. }
		\label{table:DiscussionEC}
\resizebox{\columnwidth}{!}{
\setlength{\tabcolsep}{0.6mm}
\begin{tabular}{ccc}
\toprule
U     & U+EC1 & U+EC2 \\ \midrule
27.55 & 27.23 & 27.16 \\
0.959 & 0.950 & 0.947 \\
22.99 & 22.68 & 22.59 \\
0.887 & 0.885 & 0.878 \\ \bottomrule
\end{tabular}}
\end{subtable}
\begin{subtable}[c]{0.241\textwidth}
\centering
\setlength{\abovecaptionskip}{0.05cm}
\caption{\fontsize{7.1pt}{\baselineskip}\selectfont  Additional tasks (IVIF).}
		\label{table:DiscussionIVIF}
\resizebox{\columnwidth}{!}{
\setlength{\tabcolsep}{0.6mm}
\begin{tabular}{l|cc}
\toprule
Metrics & DeRUN & DeRUN+$L_{ISIC}$ \\ \midrule
PSNR~$\uparrow$    & 17.58 & 18.03           \\
SSIM~$\uparrow$    & 0.753 & 0.760           \\
AG~$\uparrow$      & 6.98  & 7.05            \\
EN~$\uparrow$      & 7.17  & 7.24     \\ \bottomrule      
\end{tabular}}
\end{subtable}
\begin{subtable}[c]{0.204\textwidth}
\centering
\setlength{\abovecaptionskip}{0.05cm}
\caption{\fontsize{7.1pt}{\baselineskip}\selectfont  Additional tasks (SOD).}
		\label{table:DiscussionSOD}
\resizebox{\columnwidth}{!}{
\setlength{\tabcolsep}{0.6mm}
\begin{tabular}{l|cc}
\toprule
Metrics & RUN   & RUN+$L_{ISIC}$ \\ \midrule
$M$~$\downarrow$      & 0.022 & 0.021         \\
$F_\beta$~$\uparrow$       & 0.886 & 0.892         \\
$E_\phi$~$\uparrow$       & 0.953 & 0.958         \\
$S_\alpha$~$\uparrow$       & 0.916 & 0.919   \\ \bottomrule     
\end{tabular}}
\end{subtable}\vspace{-2mm}
\end{table}
\begin{figure*}[htbp!]
\setlength{\abovecaptionskip}{0cm}
	\centering
	\includegraphics[width=\linewidth]{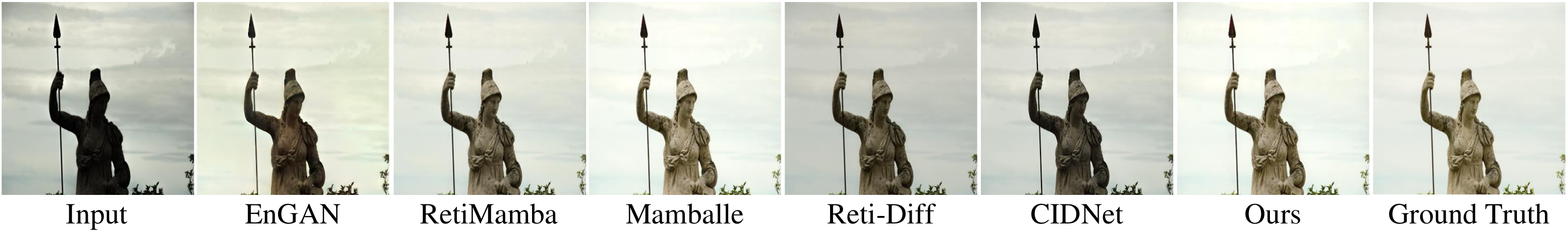} 
	\caption{Limitations.
    }
	\label{fig:limitation}
	\vspace{-3.5mm}
\end{figure*}


\section{Discussions} \label{discussion}
In addition to proposing a powerful DUN-based method, this paper thoroughly explores the intrinsic advantages of DUNs and investigates their deployment for image restoration. From our analysis, we derive several significant conclusions:

First, as shown in Table~\ref{table:DiscussionRT}, the inclusion of explicit regularization terms is particularly beneficial for DUN-based frameworks, especially lightweight variants such as UnfoldIR-t (U-t) and UnfoldIR-s (U-s). These terms introduce explicit, task-specific priors, reducing the need for networks to implicitly learn complex priors. Consequently, this strategy decreases the number of required parameters and mitigates limitations faced by lightweight networks in modeling sophisticated priors.

Second, additional network connections—referred to here as extra connections (EC)—which have been commonly employed in previous DUN-based image restoration methods, such as EC1~\cite{wu2021dense} and EC2~\cite{mou2022deep}, do not always yield performance improvements. As illustrated in Table~\ref{table:DiscussionEC}, these connections can sometimes degrade performance for two primary reasons: (1) DUNs inherently possess structured connections that rigorously derive from mathematical principles; (2) hence, adding arbitrary connections that are typically beneficial in purely learning-based, "black-box" connections can disrupt the mathematically principled structure of DUNs and impair their performance.

Third, as demonstrated in Tables~\ref{table:DiscussionIVIF} and \ref{table:DiscussionSOD}, our proposed $L_{\text{ISIC}}$ loss—a self-consistency supervision strategy that leverages the unique multi-stage structure of DUNs—shows strong generalizability to other image processing tasks. For instance, the DeRUN framework~\cite{he2023degradation} in infrared and visible image fusion (IVIF), evaluated by additional metrics such as average gradient (AG) and entropy (EN), clearly benefits from $L_{\text{ISIC}}$ (where higher AG and EN scores indicate better performance). Similarly, RUN~\cite{he2025run} in salient object detection (SOD) also demonstrates improved performance.

Finally, as presented in Table~\ref{table:DownstreamApplications}, DUNs effectively enhance downstream tasks by leveraging outputs from multiple unfolding stages. These outputs serve as enhancement-based data augmentation inputs for downstream algorithms, thereby improving their overall performance.

\section{Limitations and Future Work} \label{limitation}
As illustrated in~\cref{fig:limitation}, our method fails to recover certain subtle texture details. This limitation is common among existing approaches and is likely since such fine details, when obscured in dark regions, may be misinterpreted as artifacts or degradation and consequently removed during the restoration process.

This issue poses a challenge for preserving semantic consistency across image components and may lead to visually unnatural results. To address this, future work will explore the integration of high-level vision tasks to better capture and interpret the semantic context of image components. Additionally, we plan to enhance our DUN-based framework by incorporating generative techniques, such as diffusion models, to improve its capacity for producing results with higher perceptual fidelity.

\end{document}